\documentclass[preprint,12pt]{elsarticle}

\usepackage{graphicx}

\usepackage{amsmath, amssymb}
\usepackage{algorithm}
\usepackage{algorithmic}

\usepackage{amsthm}

\newcommand{\bmu}{\mbox{\boldmath{$\mu$}}}

\newcommand{\bnabla}{\mbox{\boldmath{$\nabla$}}}
\newcommand{\bsigma}{\mbox{\boldmath{$\sigma$}}}

\newcommand{\bK}{\mbox{\boldmath{$K$}}}

\newcommand{\bDelta}{\mbox{\boldmath{$\Delta$}}}

\newcommand{\bA}{\mbox{\boldmath{$A$}}}

\newcommand{\bI}{\mbox{\boldmath{$I$}}}

\newcommand{\bE}{\mbox{\boldmath{$E$}}}
\newcommand{\bp}{\mbox{\boldmath{$p$}}}

\newcommand{\bx}{\mbox{\boldmath{$x$}}}

\newcommand{\bX}{\mbox{\boldmath{$X$}}}

\journal{Images and Vision Computing}

\begin{document}

\begin{frontmatter}



\title{Combining contextual and local edges for line segment extraction in cluttered images}


\author{Rui F. C. Guerreiro\corref{cor1}}
\ead{ruifcguerreiro@hotmail.com}

\cortext[cor1]{Corresponding author}

\address{Institute for Systems and Robotics, Instituto Superior T\'{e}cnico, Lisboa, Portugal}

\begin{abstract}

Automatic extraction methods typically assume that line segments are pronounced, thin, few and far between, do not cross each other, and are noise and clutter-free. Since these assumptions often fail in realistic scenarios, many line segments are not detected or are fragmented. In more severe cases, {\it i.e.}, many who use the Hough Transform, extraction can fail entirely. In this paper, we propose a method that tackles these issues. Its key aspect is the combination of thresholded image derivatives obtained with filters of large and small footprints, which we denote as {\it contextual} and {\it local edges}, respectively. Contextual edges are robust to noise and we use them to select {\it valid} local edges, {\it i.e.}, local edges that are of the same type as contextual ones: dark--to--bright transition of vice-versa. If the distance between valid local edges does not exceed a maximum distance threshold, we enforce connectivity by marking them and the pixels in between as edge points. This originates connected edge maps that are robust and well localized. We use a powerful two-sample statistical test to compute contextual edges, which we introduce briefly, as they are unfamiliar to the image processing community. Finally, we present experiments that illustrate, with synthetic and real images, how our method is efficient in extracting complete segments of all lengths and widths in several situations where current methods fail.


\end{abstract}


\end{frontmatter}

\section{Introduction}
\label{sec:intro}

Line segments provide important information about the geometric content of real-life images. Since most man-made objects are made of flat surfaces, the contours needed for the interpretation of such 2D images as well as 3D world scenes often consist of line segments. Also, many complex shapes accept an economic and simple description in terms of straight lines. This has been used, {\it e.g.}, for localizing vanishing points \cite{Zhang02videocompass} or to match line segments across distinct views \cite{Schmid97automaticline}. Other applications include, {\it e.g.}, rectangle detection~\cite{rectangleDetection08}, the inference of shape from lines~\cite{shapeFromLines96}, map-to-image registration~\cite{MapToImageRegistration}, 3D reconstruction~\cite{lineDrawingTo3D}, or even image compression~\cite{Ageenko98compressionof}.


Although automatic line segment extraction has been researched actively in the past decades, current solutions make use of implicit strong assumptions that limit their applicability to simple and mostly unrealistic scenarios. Typical assumptions are that, {\it e.g.}, line segments are pronounced, thin, occur in small amounts, do not cross each other, are located away from noise or clutter. Some methods also assume that images have few data apart from line segments, such as textures or contours. Since these assumptions often fail in realistic scenarios, many line segments are not detected, or are fragmented to various extents. In more severe cases, extraction can fail altogether. For this reason, which we detail in the sequel, the robust detection of line segments in realistic scenarios remains an open frontier (see \cite{newPaperHT,LSD10,borkar2011,novelHT4enhacedAccumulatorArray11} for examples of recent advances).

\subsection{Overview of methods for line segment extraction}


The Hough transform (HT)~\cite{Hough62,DudaHart72} is the most popular method to detect lines in images. It is a likelihood-based parameter extraction technique that, basically, indicates that the largest accumulation of edge points correspond to image lines. It uses a Hough space, a two-dimensional space where each point represents a line in the image, and each edge point in the image votes on the region of the Hough space that represents the pencil of all the image lines that go through that edge point. By processing all edge points, the votes for each location in the Hough space are accumulated and the locations with larger number of votes correspond to the most likely parameterizations of the lines in the image. Later, the HT was extended for extracting line segments. After obtaining the line parameterizations with the usual HT, the start and end points of the segments are obtained using a gap-and-length method~\cite{gapAndLength05}, the shape of the spread of votes in the Hough space~\cite{butterflyHough98,novelHT4enhacedAccumulatorArray11} or extra accumulators~\cite{ies_2011_teutsch_ipcv}.


The success of the HT comes from its global nature, since all points in a line contribute to its detection --- in fact, it was proven that it implements a statistically robust estimator for finding lines~\cite{HTEstimator04}. It has, however, three major issues when used to extract line segments in complex images. Firstly, it requires an edge detection scheme such as, {\it e.g.}, the Canny edge detector~\cite{canny86}, to generate its input edge map. Edge detection is by itself a hard problem, recognized as ill-posed in general~\cite{ill-posedproblems88}, where delicate balances occur between edge localization and noise reduction, and detecting spurious edge points in noisy or textured areas and missing faint edges. As a consequence, edge detectors typically make use of small (local) filters and high thresholds for accepting an edge point, resulting in partial or complete segment mis-detections. This counteracts the global nature of the HT.


Secondly, since all votes originated by an edge point are {\it wrong} except for the one corresponding to the actual segment, the amount of noise in the Hough space is significant for images with many edge points~\cite{HTValidity98}. This makes it difficult to identify the most likely parameterization of actual lines in the Hough space. This is particularly critical for short segments, since they originate small peaks that are hard to identify~\cite{FastHoughTransformMore95}. Accidental alignments of unrelated edge points can also originate false peak detections~\cite{HTValidity98,ourTIPHTPaper12}. An example where Hough space contamination due to poor edge detection and erroneous vote accumulations leads to complete extraction breakdown is shown in~\cite{ourTIPHTPaper12}. Although Duda and Hart~\cite{DudaHart72} argued as early as 1972 that the noise in the Hough space can be reduced by taking connectivity between collinear points into account, this topic received little attention in the past (exceptions are~\cite{ConnectiveHT93,HTmodifiedLineConnectivityLineThickness97,HTValidity98,ourTIPHTPaper12}).


Thirdly, the way in which the HT was adapted to extract line segments, {\it i.e.}, by taking the output of the HT and obtaining the start and end points of the segments, originates issues of its own. While initially each point in the Hough space accumulated votes supporting the existence of only one image line, now the votes can refer to multiple collinear line segments of various lengths and different start and end points. Since only the number of collinear edge points is stored in a conventional Hough space, the votes of individual line segments cannot be distinguished. This means that a local maximum in the Hough space no longer implies a maximum likelihood that a line or line segments actually exist in the image with such parameterization --- therefore, this adaptation of the HT is not a statistically robust line segment extractor. Few papers deal with this topic as well, as in~\cite{HTValidity98,ourTIPHTPaper12}.


Another issue of the HT is that it cannot deal properly with wide line segments~\cite{HTmodifiedLineConnectivityLineThickness97}, {\it i.e.}, segments made up of blurred edges, since the highest number of votes corresponds to the diagonal of the segment rather than the segment itself. The limitations of the HT in handling complex images have been pointed out by several authors, {\it e.g.}, \cite{FastHoughTransformMore95,LSD10,ourTIPHTPaper12}, and many efforts have been made to alleviate its problems. They include, {\it e.g.}, the use of the edge direction to reduce the accumulation of spurious votes in the Hough space~\cite{HoughSurvey87,ExtractLinesNoisyImageUsingDirectionalInfo06}, the sequential processing and removal of the strongest peaks in the Hough space \cite{FastHoughTransformMore95,ies_2011_teutsch_ipcv}, sub-sampling of the edge map (randomized HT)~\cite{probalisticAndNonProbabilistic95}. Other authors addressed storage and computational issues of the HT by proposing a hierarchical scheme \cite{FastHoughTransform86}, multiple accumulator resolutions \cite{AdaptiveHoughTransform87} and a probabilistic formulation~\cite{probabilisticHT91}. The thickness of line segments and edge point connectivity is used in~\cite{HTmodifiedLineConnectivityLineThickness97} to change the value of each vote in a standard HT. However, none of these methods tackle the fundamental issues of the HT in extracting line segments in complex images.

The other set of popular methods for line segment extraction can be categorized as {\it local methods}, due to their reliance on local decisions rather than global ones (see \cite{Nevatia1980257,extractingStraightLines86,Guru2004,comparisonLineExtraction07,LSD10} for examples). The majority of local methods use three steps: first, they detect edge points and chain them (using, {\it e.g.}, the method in~\cite{edgeChaining92}); then, a rough estimate of the segment direction is computed using, {\it e.g.}, total least-squares regression~\cite{comparisonLineExtraction07}; and finally, they refine and extend the segment by including new edge points that approximately fit the line. The final step usually involves alternating between two stages until convergence~\cite{comparisonLineExtraction07}: inclusion of new edge points that are close to the candidate line, according to a distance measure; and re-estimation of the line segment parameters from the new set of edge points. In realistic scenarios, the initial step of detecting edge points and large connected regions belonging to a single segment is hard due to texture, low-contrast regions, crossing segments, and noise. The resulting smaller edge point chains hurt the reliability of the regression step and, finally, as it is typical with this type of alternating methods, a poor initial model for the line segment model may compromise the final refinement and extension results. Variations of the basic method include, {\it e.g.}, \cite{edgeChainLineCutting92}, which takes the chained edge points and cuts them into line segments, using a straightness criterion. References~\cite{Guru2004,directRegression97} and~\cite{twoFrame03} bypass the chaining of edge points and fit a line directly to all edge points inside a sliding window. The segment direction is estimated roughly in~\cite{featureExtraction78} using a so-called local HT and taking the peaks of local orientation histograms, computed at each edge point. Two popular local methods for line segment detection are \cite{extractingStraightLines86} and the LSD (Line Segment Detector) of \cite{LSD10}. The method in \cite{extractingStraightLines86} coarsely quantizes the local orientation angles, chains adjacent pixels with identical orientation labels, and fits a line segment to the grouped pixels. LSD \cite{LSD10} extends this idea by using continuous angles and eliminates false line segment detections by using the Helmholtz principle of \cite{LSDAllCombinations06}. These methods result computationally simple but lack robustness to deal with the imperfections that occur in realistic scenarios.


In~\cite{ourTIPHTPaper12}, we propose a better adaptation of the HT to the extraction of line segments, which retains its global aspect and solves its main issues in dealing with complex images. It starts by computing image derivatives using small filters and obtaining the directions for which there is a predominance of positive or negative derivatives (corresponding to a dark-to-bright transition in the image or vice-versa). Such predominance gives a rough indication that a line segment might be present at that location (it implements the sign test, as we discuss later) and only edge points within areas of strong predominance are allowed to contribute. Then, instead of using a single Hough space, where collinear line segments are indiscernible, we use a local HT for each line segment. Connectivity is incorporated in the voting process, by only accounting for the contributions of edge points whose position and directional content agree with potential line segments. As a consequence, the vast majority of spurious votes are eliminated and the peaks in the local Hough spaces correspond to line segments of maximum length. Unfortunately, the computational complexity of the proposed method is prohibitive for many applications (computation times of 100 to 1000 seconds are reported in~\cite{ourTIPHTPaper12}) and thick transitions are not obtained correctly, as in the standard HT. We now believe that the requirements of global methods, of storing comprehensive data in order to eliminate early decisions that compromise robustness, make such methods computationally too complex for most applications.

\subsection{Proposed approach}


The novel and key aspect of our approach is the combination of what we denote {\it contextual} and {\it local edges}. Contextual edges are thresholded image derivatives obtained with filters of large footprint, which reduce the influence of noise when identifying image transitions. Although large footprint filters are robust to noise, edge localization is imprecise since every transition is smeared by their large point spread function. On the other hand, local edges are thresholded image derivatives obtained with filters of very small footprint, {\it e.g.}, Sobel, Prewitt, and Roberts operators. Since these filters are small, edges are localized precisely but noise may originate erroneous detections of multiple directions and amplitudes. Our proposal combines contextual and local edges obtained at the same pixels by taking the sign of the contextual edge and using it to identify so-called valid local edges with the same sign. The edge sign indicates if it is a dark--to--bright transition or the opposite.


In complex images, valid local edges are disconnected with each other, due to noise and clutter. To obtain connected edge maps, we handle connectivity explicitly in the combination process by checking if the valid local edges, along a given direction, are at a distance not greater than a maximum distance threshold from each other. If so, the pixels corresponding to valid local edges and those in between are marked as edge points. The resulting edge detector has the robustness of contextual edges in dealing with noise and the localization of local edges, as idealized by Canny~\cite{canny86}. Since the combination of contextual and local edges originates pixel-thin (connected) regions of the length of the line segment, a simple region growing and rectangle fitting methodology (which we detail later) suffices for the individual thin segments to be combined into line segments of all lengths and widths. 

Typical contextual edge detectors handle noise by applying a low-pass filter of large footprint such as, {\it e.g.}, Gabor or steerable filters~\cite{Freeman91thedesign}, followed by a derivation step and binarization with a fixed threshold. However, since other unwanted high-frequency variations due to, {\it e.g.}, textures and clutter from interfering line segments and non-rectilinear image data, are common in complex images and have unpredictable amplitudes --- unlike typical image white noise, which is constant throughout the image --- an adaptive threshold is beneficial. For this purpose, we take both the mean and the variance of the two sets of pixels and use a two-sample statistical test to determine if a contextual edge exists. We consider that each set of pixels follows a Normal distribution with the sampled parameters, compute the Total Variation distance between them and threshold the confidence interval for the null hypothesis that both distributions are actually the same.


By using large (yet limited) contextual filters, our semi-global method results simple. Also, by doing away with the issues of local edges in a more effective way than typical global methods, through a combination with contextual edges, it results typically more robust as well in obtaining line segments of all lengths and transition widths. This is demonstrated in the experimental section, where we present a complexity analysis and illustrative results using synthetic and real images to compare our method with other methods: the standard HT \cite{DudaHart72}, the state-of-the-art of local methods LSD \cite{LSD10}, and our previous HT-based method~\cite{ourTIPHTPaper12}.

\subsection{Paper organization}

The organization of the remaining of the paper is as follows. In section~\ref{sec:statisticalEdgeDetection} we provide a brief introduction to statistical edge detectors and detail our implementation. Section~\ref{sec:edgeDetection} details the combination of contextual and local edges, including the explicit tackling of connectivity. Rectangle fitting is described in~\ref{sec:recFittingValidation} and the experimental results are reported in Section~\ref{sec:experiments}. Section~\ref{sec:conclusions} concludes the paper.

\section{Statistical edge detection}
\label{sec:statisticalEdgeDetection}

\begin{figure}[hbt!]
\centerline{
\includegraphics[width=.99\linewidth]{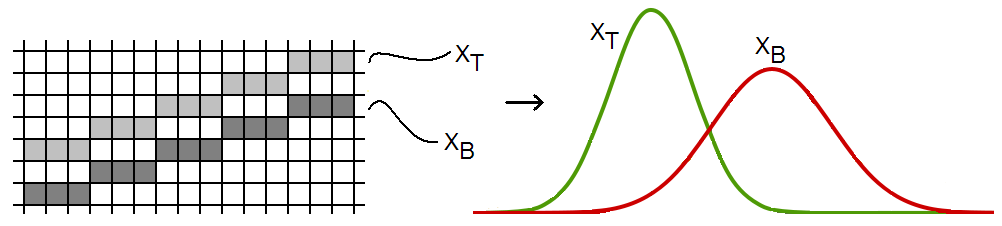}
}
\caption{Illustration of two-sample tests. The pixels (left) are samples of underlying distributions $\bX_T$ and $\bX_B$ (right). The two-sample test determines if they are the same.}\label{fig:footprintOfOrientedFeatureDetectors}
\end{figure}

Low-pass filtering prior to derivation is used extensively in edge detection schemes to reduce the effect of noise, albeit with small footprints for small localization error. This can be seen roughly as computing the mean of two sets of $M$ pixels, comparing them and binarizing the result using threshold $C$. In this scheme, the value of $C$ is critical, since a small $C$ originates erroneous edges, and a large $C$ misses them. In simpler scenarios where noise has a constant variance $\sigma^2$ throughout the image, the optimal threshold between two sets of pixels can be defined optimally for a given confidence interval ({\it e.g.}, the standard error of the mean, $C = \sigma/\sqrt{M}$). However, since complex images contain unwanted high-frequency variations of unpredictable variance such as, {\it e.g.}, textures and clutter from other image data, a fixed threshold is far from optimal. This suggests that a more comprehensive statistical analysis is able to improve edge detection. Such studies were initiated by Bovik et al.~\cite{Bovik86}, which used two-sample statistical tests in the context of edge detection. Two-sample tests take two sets of pixel values, illustrated in Fig.~\ref{fig:footprintOfOrientedFeatureDetectors}, and considers that they are samples of two underlying distributions, $\bX_T$ and $\bX_B$. The test then checks the null hypothesis $\mathcal{H}_0$ that the underlying probability distributions are in fact the same~\cite{twoSampleTests}. A contextual edge exists between the samples of $\bX_T$ and $\bX_B$ if the distributions are deemed different.

\subsection{Typical approaches}

The parameters of each distribution can be estimated from the samples by computing $\hat{\mu}_i = \frac{1}{M}\sum_{j = 1}^{M} \bx_{ij}$ and $\hat{\sigma}_i^2 = \frac{1}{M-1}\sum_{j = 1}^{M} \left(\bx_{ij}-\hat{\mu}_i\right)^2$ (with $i \in \{T, B\}$). If the two sets of samples are assumed to be Normally distributed, $\bX_T \sim \mathcal{N}(\mu_T,\sigma_T^2)$ and $\bX_B \sim \mathcal{N}(\mu_B,\sigma_B^2)$, various criteria can then be used. 

For the $t$-test, the distributions are the same if their means coincide. Since the sample mean varies with the sample variance and the number of samples through formula $\sigma_{\hat{\mu}_i} = \hat{\sigma}_i/\sqrt{M}$, the $t$-test is given by
\begin{equation}
t = \left(\hat{\mu}_T -\hat{\mu}_B\right)/\sqrt{\frac{\hat{\sigma}_T^2+\hat{\sigma}_B^2}{M}}~\text{\cite{twoSampleTests}}.\label{eq:tTest}
\end{equation}
Once a $t$-value is determined, the probability that the test statistic would take a value at least as extreme as the one observed, denoted $p$-value, is computed, for $M - 1$ degrees of freedom. If the calculated $p$-value is below the threshold chosen for statistical significance (usually the 0.10, the 0.05, or 0.01 level), the distributions are deemed different (two-tailed test) or larger than the other (one-tailed test)~\cite{twoSampleTests}.


If no assumption is made about the distributions of $\bX_T$ and $\bX_B$, nonparametric tests can be used instead. These tests are more general and robust to outliers~\cite{twoSampleTests} but also computationally more intensive, since they require expensive sorting operations. Some tests compute empirical cumulative distributions (ecd) from samples $\bX_T$ and $\bX_B$ and then compare them using some distance measure. The Kolmogorov--Smirnov test obtains the maximum distance between the ecds and the Fisz--Cram\'er--Von Mises test integrates the squared difference between the ecds. The Wilcoxon Mann--Whitney is a popular rank order test that mixes the samples of both distributions, sorts and ranks them. The difference between the distributions is assessed by adding the ranks of one distribution and comparing with the added ranks of the other~\cite{twoSampleTests}.

The success of two-sample statistical tests in contextual edge detection in noisy images is shown in~\cite{comparisonStatisticalTests02}. Reference~\cite{FesharakiHellestrand94} uses a $t$-test for detecting edges and a mixture of Normal distributions models noisy data in the edge detector of~\cite{Thune97}. Various statistical tests are used in the neural network approach~\cite{statisticalNN06}. In our previous line segment extraction method~\cite{ourTIPHTPaper12}, the number of local positive image derivatives minus the number of negative ones is used to assess the predominance of dark-to-bright transitions or vice-versa, indicating a high chance of having found a line segment. This procedure implements the sign test, a nonparametric paired test that tests the null hypothesis that there is "no difference in medians" between $\bX_T$ and $\bX_B$. Despite lacking the statistical power of other tests, it has very general applicability, as it makes very few assumptions about the nature of the distributions under test~\cite{twoSampleTests}.


\subsection{Our approach}


Since nonparametric tests are computationally expensive, we use a parametric test instead. Although the $t$-test is arguably the most frequently used parametric test, it assumes that coinciding mean values imply identical distributions (this is clear in equation~\eqref{eq:tTest}), which is not coherent with the intuition that pixel distributions with the same sample mean but different sample variances can be deemed visually different. To enable this feature, we use the Total Variation (TV) distance between two Normal distributions~\cite{totalVariationBook},
\begin{align}
\delta(\bX_T,\bX_B) &= \frac{1}{2}\displaystyle\int_{-\infty}^{\infty} \left| f(\xi;\hat{\mu}_T,\hat{\sigma}_T^2) - f(\xi;\hat{\mu}_B,\hat{\sigma}_B^2)\right| d\xi \in [0,1],\label{eq:TVdistanceBasic}
\end{align}
where $f(\cdot; \mu, \sigma^2)$ represents the probability density function (pdf) of the Normal distribution. The TV distance outputs the integral of the (linear) distance between the pdfs of $\bX_T$ and $\bX_B$. As we show below, the linear aspect assures that only the ratio between the sample standard deviations is taken into account, instead of their actual values. Although more comprehensive tests are needed, the higher sensitivity and robustness of the TV distance in detecting the perceived line segments, in contrast with non-linear distances such as the Kullback-Leibler and Hellinger divergences, lead us to believe that this feature of the TV distance best emulates the human visual system in estimating the boundary strength between different areas.



\begin{figure}[hbt!]
\centerline{
\includegraphics[width=.65\linewidth]{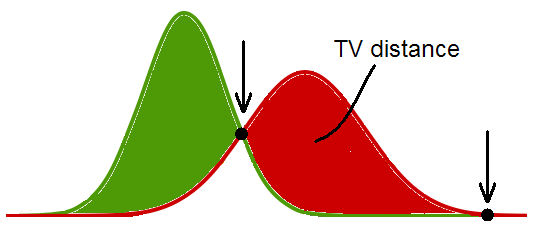}
}
\caption{Illustration of distribution pdf intersection points and area for TV distance.}\label{fig:TVdistance}
\end{figure}

To simplify the calculation of~\eqref{eq:TVdistanceBasic}, let $\xi_i \in \{-\infty,\hat{\xi}_1,\hat{\xi}_2,\infty\}$ be the points in which the pdf of distributions $\bX_T$ and $\bX_B$ are equal, as illustrated in Fig.~\ref{fig:TVdistance}. Points $\xi_i$ (with $\xi_i \leq \xi_{i+1}$, $\forall i \in \{1,\hdots,3\}$) determine where one pdf becomes larger than the other and, thus, helps in dealing with the magnitude operator and enables the use of the cumulative density function of the Normal distribution, $F(\cdot; \mu, \sigma^2)$,
\begin{align}
\delta(\bX_T,\bX_B) &= \frac{1}{2}\displaystyle \sum_{i = 1}^{3} \displaystyle \int_{\xi_i}^{\xi_{i+1}} \left| f(\xi;\hat{\mu}_T,\hat{\sigma}_T^2) - f(\xi;\hat{\mu}_B,\hat{\sigma}_B^2)\right| d\xi\notag
\\
&= \frac{1}{2}\displaystyle \sum_{i = 1}^{3} \left| \displaystyle \int_{\xi_i}^{\xi_{i+1}} f(\xi;\hat{\mu}_T,\hat{\sigma}_T^2) - f(\xi;\hat{\mu}_B,\hat{\sigma}_B^2) d\xi \right|\label{eq:TVdistanceFinal}
\\
& = \frac{1}{2}\displaystyle \sum_{i = 1}^{3} \left|F(\xi_{i+1};\hat{\mu}_T,\hat{\sigma}_T^2)-F(\xi_i;\hat{\mu}_T,\hat{\sigma}_T^2)-F(\xi_{i+1};\hat{\mu}_B,\hat{\sigma}_B^2)+F(\xi_i;\hat{\mu}_B,\hat{\sigma}_B^2)\right|,\notag
\end{align}
To determine $\hat{\xi}_1$ and $\hat{\xi}_2$, we make $f(\hat{\xi};\hat{\mu}_T,\hat{\sigma}_T^2) = f(\hat{\xi};\hat{\mu}_B,\hat{\sigma}_B^2)$, which results in $\hat{\xi} = \left(-b \pm \sqrt{b^2-4ac}\right)/2a$, where $a=1/\left(2\hat{\sigma}_T^2\right)-1/\left(2\hat{\sigma}_B^2\right)$, $b = -\hat{\mu}_T/\left(\hat{\sigma}_T^2\right)+\hat{\mu}_B/\left(\hat{\sigma}_B^2\right)$, and $c = \hat{\mu}_T/\left(2\hat{\sigma}_T^2\right)-\hat{\mu}_B/\left(2\hat{\sigma}_B^2\right)-\ln\left(\hat{\sigma}_B/\hat{\sigma}_T\right)$. If all parameters are identical, then $\delta(\bX_T,\bX_B) = 0$. If only the sample variances are identical, the pdf of both distributions cross at only one point, $\hat{\xi}_1 = \hat{\xi}_2 = \left(\hat{\mu}_T-\hat{\mu}_B\right)/2$. For different sample variances, two crossing points are obtained.

To eliminate redundant calculations in the remainder of our method, we compute a look-up table with relevant TV distances. Since the TV distance requires four parameters, $(\hat{\mu}_T, \hat{\sigma}_T, \hat{\mu}_B, \hat{\sigma}_B)$, it seems that a four-dimensional look-up table is needed, which is expensive to compute and store. However, since we are computing the linear distance between two distributions, it is not the actual values of $\hat{\mu}_T$ and $\hat{\mu}_B$ that are needed but how they relate with each other. Similarly, the actual values of $\hat{\sigma}_T$ and $\hat{\sigma}_B$ are not needed but only the relation between them as well. Let variables $\hat{\mu}'_B$ and $\hat{\sigma}_B'$ be
\begin{equation}
\begin{array}{cc}
\hat{\mu}'_B = \frac{|\hat{\mu}_B - \hat{\mu}_T|}{\min(\hat{\sigma}_T,\hat{\sigma}_B)} &
\hat{\sigma}_B' = \frac{\max(\hat{\sigma}_T,\hat{\sigma}_B)}{\min(\hat{\sigma}_T,\hat{\sigma}_B)}
\end{array}.\label{eq:normalization4LUT}
\end{equation}
Assuming that $\hat{\mu}_T \leq \hat{\mu}_B$ and $\hat{\sigma}_T \leq \hat{\sigma}_B$, equations~\eqref{eq:normalization4LUT} become
\begin{equation}
\begin{array}{cc}
\hat{\mu}_B = \hat{\mu}_T + \hat{\mu}'_B\hat{\sigma}_T & \hat{\sigma}_B = \hat{\sigma}_T\hat{\sigma}'_B.
\end{array}\label{eq:normalizationOldVar}
\end{equation}
Replacing~\eqref{eq:normalizationOldVar} in~\eqref{eq:TVdistanceBasic} and substituting variable $\zeta = \left(\xi - \mu_T\right)/\sigma_T$, we have
\begin{align}
\delta(\bX_T,\bX_B) &= \frac{1}{2}\displaystyle\int_{-\infty}^{\infty} \left| f(\xi;\hat{\mu}_T,\hat{\sigma}_T^2) - f(\xi;\hat{\mu}_T + \hat{\mu}'_B\hat{\sigma}_T, \hat{\sigma}_T^2\hat{\sigma}_B^{'2})\right| d\xi\notag
\\
&= \frac{1}{2}\displaystyle\int_{-\infty}^{\infty} \left| f(\zeta;0,1)/\hat{\sigma}_T - f(\zeta;\hat{\mu}'_B, \hat{\sigma}_B^{'2})/\hat{\sigma}_T\right| \hat{\sigma}_T d\zeta\label{eq:TVdistanceBasicNorm}
\\
&= \frac{1}{2}\displaystyle\int_{-\infty}^{\infty} \left| f(\zeta;0,1) - f(\zeta;\hat{\mu}'_B, \hat{\sigma}_B^{'2})\right| d\zeta,\notag
\end{align}
which shows that the TV distance depends only on two parameters (analogous demonstration show the same result in any scenario). A two-dimensional look-up table $\bDelta(\hat{\mu}'_B,\hat{\sigma}'_B)$ is then filled in for various $(\hat{\mu}'_B,\hat{\sigma}'_B)$, with $\hat{\mu}_B \geq 0$ and $\hat{\sigma}'_B \geq 1$, using~\eqref{eq:TVdistanceFinal}. During normal operation, the parameters are normalized using equations~\eqref{eq:normalization4LUT} and the corresponding TV distance is accessed.


\section{Combining contextual and local edges using connectivity}
\label{sec:edgeDetection}

In this section, we compute contextual edges and combine them with local ones to form connected edge maps. Since samples $\bX_T$ and $\bX_B$, the inputs of the two-sample test at the heart of contextual edge extraction, are taken from long and thin footprints (as illustrated in Fig.~\ref{fig:footprintOfOrientedFeatureDetectors}), each test is effective only in a small angular range. Therefore, contextual edges must be computed along multiple directions to span the entire $180^{\circ}$ range. For every direction, we use a running average approach to take samples $\bX$ at each point and, assuming that they follow a Normal distribution, obtain their parameters. Then, we take the parameters refering to $\bX_T$ and $\bX_B$, compute the Total Variation distance detailed in the previous chapter and describe each pixel (and direction) as containing a {\it negative}, a {\it positive} or {\it no contextual edge}, using quantization thresholds defined later. If there is a contextual edge, it is used to select {\it valid} local edges. Local edges consist of image derivatives obtained by convolving the image with kernels of very small footprint, central difference kernels, whose results are also described as being a negative, a positive or no local edge. Valid local edges then consist of local edges that have the same sign as the contextual one. If the distance between valid local edges does not exceed a maximum distance threshold $d$, those edges and the pixels between them are marked as edge points.

\subsection{Computing Normal distribution parameters}
\label{sec:computingNormalDistributionParameters}

\begin{figure}[htb]
\centerline{
\includegraphics[width=0.85\linewidth]{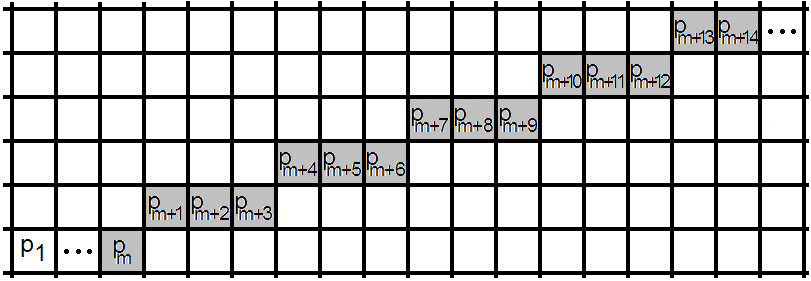}
}
\caption{Illustration of $M = 15$ pixels starting at pixel $\bp_m$ along direction $\theta_n$.}\label{fig:lineExample}
\end{figure}

The parameters of the Normal distribution are computed for each set of samples starting at pixel $\bp_m$, illustrated in Fig.~\ref{fig:lineExample}. We use $M = 15$ pixels at each set of samples and we verified theoretical and experimentally that $N = 32$ different uniformly spaced directions leads simultaneously to neglectable angular coverage errors and good computational performance (since every pixel in the perimeter of a semicircular window of radius $M$ is used only when $N \geq \pi (M-1)$, $N = 44$ different directions should be used for $M = 15$ pixels and perfect angular performance).


\begin{figure}[htb]
\centerline{
\includegraphics[width=0.3\linewidth]{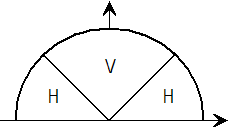}
\includegraphics[width=0.3\linewidth]{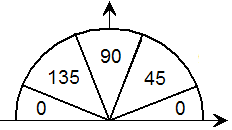}
}
\caption{Illustration of the half of a semicircle, $H(\theta_n)$ (left), and quadrant, $Q(\theta_n)$ (right).}\label{fig:edgeQuadrants}
\end{figure}

Each direction $n \in \{1,\hdots,N\}$ corresponds to angle $\theta_n = 180^{\circ}(n-1)/N$, which lies in the horizontal (H) or vertical (V) half of the semicircle $H(\theta_n)$, and in quadrant $Q(\theta_n)$ of the semicircle, both illustrated in Fig.~\ref{fig:edgeQuadrants},


\begin{align}
H(\theta_n) &= \left\{\begin{array}{ll}\text{V} & \theta_n \in [45^{\circ},135^{\circ}[ \\ \text{H} & otherwise\end{array}\right.,
\\
Q(\theta_n) &= \left\{\begin{array}{ll}0 & \theta_n \in [0^{\circ},22.5^{\circ}[ \cup [157.5^{\circ},180^{\circ}[ \\ 45 & \theta_n \in [22.5^{\circ},67.5^{\circ}[ \\ 90 & \theta_n \in [67.5^{\circ},112.5^{\circ}[  \\ 135 & \theta_n \in [112.5^{\circ},157.5^{\circ}[\end{array}\right..\notag
\end{align}

Our method works by selecting a direction $\theta_n$ and dividing the image into lines along that direction, as illustrated in Fig.~\ref{fig:lineExample}. For each point $\bp_m$ in a line, the parameters of the Normal distribution are computed by taking $M$ consecutive pixels, from pixel $\bp_m$ to pixel $\bp_{m+M-1}$, and computing their sample average and variance, $\hat{\bmu}(\bp_m)$ and $\hat{\bsigma}^2(\bp_m)$. Since the set of $M$ consecutive pixels needed for the next point in the line, $\bp_{m+1}$, is the same as the set of points for $\bp_m$ except for the points at the start and end of the set, we use a recursion that simplifies calculations and is used often: the running average. To formalize this idea, for image $\bI \in \mathbb{R}^{S_x\times S_y}$, we define a mapping that addresses each line along direction $\theta_n$,
\begin{equation}
\bA(x,y,\theta_n) = \left\{\begin{array}{ll} (x,y+\left[x\tan\theta_n\right] + \gamma S_y) & \text{if } H(\theta_n) = H
\\ (x+\left[y\tan^{-1}\theta_n\right] + \gamma S_x,y) &  \text{if } H(\theta_n) = V\end{array}\right.,\label{eq:addressPixelsInLine}
\end{equation}
where $\left[\cdot\right]$ refers to the rounding operation and integer parameter $\gamma$ is chosen so that $\bA(x,y,\theta_n)$ lies inside the image limits. If $H(\theta_n) = H$, the $y$-th line is made up of the pixels addressed by $\bA(x,y,\theta_n)$, with $x\in \{1,\hdots,S_x\}$ and in an increasing order. Analogously, if $H(\theta_n) = V$, the $x$-th line is addressed by $\bA(x,y,\theta_n)$, with increasing $y\in \{1,\hdots,S_y\}$. To compute the average and standard deviation in a recursive way, we define linear and quadratic accumulators, $\Phi_x$ and $\Phi_{x^2}$, respectively, which are updated using
\begin{align}
\Phi_x &\leftarrow \Phi_x - \bI(\bp_{m-1}) + \bI(\bp_{m+M-1})
\\
\Phi_{x^2} &\leftarrow \Phi_{x^2} - \bI^2(\bp_{m-1}) + \bI^2(\bp_{m+M-1}),\notag
\end{align}
where $\bp_{m}$ locates the current pixel, $\bp_{m-1}$ locates the pixel that is exiting the set of $M$ pixels, and $\bp_{m+M-1}$ locates the pixel that is entering the set,
\begin{equation}
\begin{array}{lcc} 
\bp_m = \bA(x,y,\theta_n), & &
\\
\bp_{m-1} = \bA(x-1,y,\theta_n), & \bp_{m+M-1} =\bA(x+M-1,y,\theta_n) & \text{if } H(\theta_n) = H
\\
\bp_{m-1} = \bA(x,y-1,\theta_n), & \bp_{m+M-1} =\bA(x,y+M-1,\theta_n) & \text{if } H(\theta_n) = V
\end{array}.
\end{equation}
The parameters of the Normal distribution, for direction $\theta_n$, are then
\begin{align}
\hat{\bmu}(\bp_m)& = \frac{1}{M}\Phi_x
\\
\hat{\bsigma}^2(\bp_m)& = \frac{1}{M-1}\Phi_{x^2}-\frac{1}{M(M-1)}\Phi_x^2,\notag
\end{align}
for each point in the image.

\subsection{Combining contextual and local edges using connectivity}
\label{sec:combContextLocal}

To compute local edges, we first capture the local directional content of image $\bI$ by computing its derivatives, through the convolution with four oriented kernels,
\begin{equation}
\bnabla_{\!\theta} \bI= \bI * \bK_{\!\theta}\,,\quad\theta\in\{0^{\circ}, 45^{\circ}, 90^{\circ}, 135^{\circ}\}\,.\label{eq:og}
\end{equation}
Local orientation has been exploited before and captured by using several types of kernels, see, {\it e.g.}, \cite{extractingStraightLines86,HoughSurvey87,LSD10}. Although kernels with a large support would smooth the noise, we use the same approach as in our previous work~\cite{ourTIPHTPaper12} and employ simple central difference kernels, since they enable more precise edge localization, by minimizing the influence of surrounding pixels. Thus, we set
\begin{equation}
\bK_{0}\!=\!\!\left[
          \begin{array}{ccc}
            \!\!\! 0 \! & \! 1 \! & \! 0 \!\!\! \\
            \!\!\! 0 \! & \! 0 \! & \! 0 \!\!\! \\
            \!\!\! 0 \! & \! -1 \! & \! 0 \!\!\!
          \end{array}
        \right]\!\!,
\bK_{45}\!=\!\!\left[
          \begin{array}{ccc}
            \!\!\! 1 \! & \! 0 \! & \! 0 \!\!\! \\
            \!\!\! 0 \! & \! 0 \! & \! 0 \!\!\! \\
            \!\!\! 0 \! & \! 0 \! & \! -1 \!\!\!
          \end{array}
        \right]\!\!,
\bK_{90}\!=\!\!\left[
          \begin{array}{ccc}
            \!\!\! 0 \! & \! 0 \! & \! 0 \!\!\! \\
            \!\!\! 1 \! & \! 0 \! & \! -1 \!\!\! \\
            \!\!\! 0 \! & \! 0 \! & \! 0 \!\!\!
          \end{array}
        \right]\!\!,
\bK_{135}\!=\!\!\left[
          \begin{array}{ccc}
            \!\!\! 0 \! & \! 0 \! & \! -1 \!\!\! \\
            \!\!\! 0 \! & \! 0 \! & \! 0 \!\!\! \\
            \!\!\! 1 \! & \! 0 \! & \! 0 \!\!\!
          \end{array}
        \right]\!.
\end{equation}

The combination of contextual and local edges occurs in three steps and is described in detail in the sequel. To summarize, in the first step, the parameters of the Normal distributions, for all pixels and directions (computed in~\ref{sec:computingNormalDistributionParameters}), and the local image derivatives (computed above) are used to compute contextual and local edges at each pixel. This occurs for each direction $\theta_n$ and progressively along each line, and the goal is to find a contextual and local edge with matching sign. Once both edge types are matched, at pixel $\bp_m$, a second step then checks if the $M$ pixels between $\bp_m$ and $\bp_{m+M-1}$ contain valid local edges that are sufficiently connected with each other, as should occur in a line segment. We consider that valid local edges are connected if they are not separated by more than a maximum distance threshold $d$. If step 2 is successful, the beginning of a line segment was found. A third step marks those pixels as {\it connected edge points} and progressively checks and marks the subsequent pixels along direction $\theta_n$ which preserve the contextual edge sign and whose valid local edges continue to be sufficiently connected. The output of this process is a set of connected edge points for each direction $\theta_n$, which is used in section~\ref{sec:recFittingValidation} to fit rectangles.


\subsubsection{Search for initial edge}


In this step, the pixels along direction $\theta_n$ are scanned to find a contextual and local edge whose spatial location and sign coincides. For $H(\theta_n) = H$ (the adaptation to $H(\theta_n) = V$ is analogous), each $y$-th line is scanned at a time ($y \in \{1,\hdots,S_y\}$) and, within each line, the pixels are scanned for all $x \in \{1,\hdots,S_x\}$ in increasing order, resulting in coordinate $\bp_1 = \bA(x,y,\theta_n)$ (to simplify the explanation of our method, we define that $\bp_m$ refers to the $m$-th pixel of the set of $M$ pixels currently being tested).

\begin{figure}[hbt!]
\centerline{
\includegraphics[width=.99\linewidth]{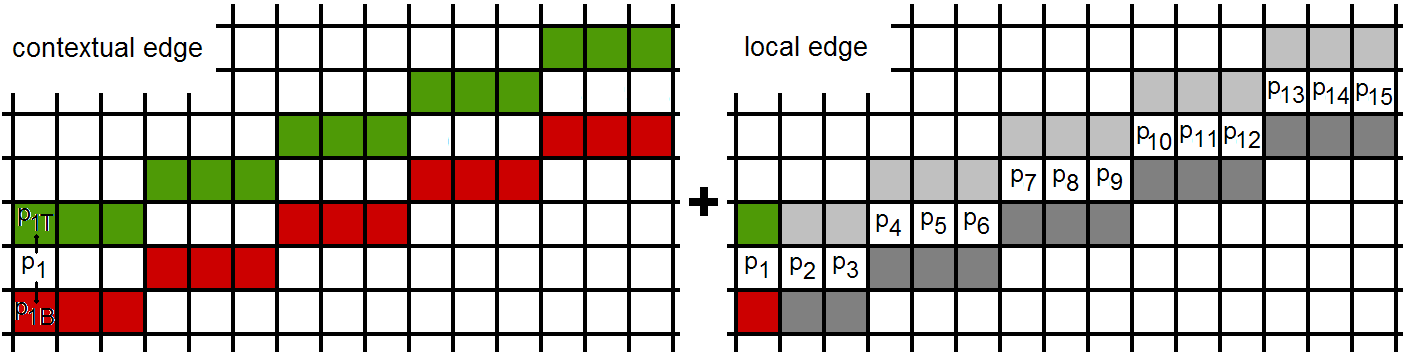}
}
\caption{Search for initial contextual (left) and valid local edge (right).}\label{fig:threePhases1}
\end{figure}

To determine if there is a contextual edge at point $\bp_1$ along direction $\theta_n$, we start by computing the coordinates of the parameters above and below $\bp_1$ (see Fig.~\ref{fig:threePhases1} for an illustration), $\bp_{1_T} = \bA(x-\left[\sin\theta_n\right],y-\left[\cos\theta_n\right],\theta_n)$ and $\bp_{1_B} = \bA(x+\left[\sin\theta_n\right],y+\left[\cos\theta_n\right],\theta_n)$, and obtain the parameters of the two distributions, $\hat{\mu}_T = \hat{\bmu}(\bp_{1_T})$,  $\hat{\sigma}_T^2 = \hat{\bsigma}^2(\bp_{1_T})$, $\hat{\mu}_B = \hat{\bmu}(\bp_{1_B})$, $\hat{\sigma}_B^2 = \hat{\bsigma}^2(\bp_{1_B})$. The parameters are normalized using equation~\eqref{eq:normalization4LUT} and the two-dimensional look-up table containing the TV distances is accessed, $\bDelta(\mu'_B,\sigma'_B)$. Since the type of transition is important for line segment extraction, {\it i.e.}, if it is a dark-to-light transition or vice-versa, contextual and local edges should include this information. For this effect, the contextual edge is given by $\delta_{{X_T}{X_B}} = \bDelta(\mu'_B,\sigma'_B)\text{sgn}(\hat{\mu}_T-\hat{\mu}_B)$. A contextual edge exists if $|\delta_{{X_T}{X_B}}| \geq C$.

There is a local edge at pixel $\bp_1$ if $|\bnabla_{\!Q(\theta_n)}(\bp_1)| \geq L_C$. If it has the same sign as the contextual one ({\it i.e.}, if condition $\text{sgn}\left(\delta_{{X_T}{X_B}}\right) = \text{sgn}\left(\bnabla_{\!Q(\theta_n)}(\bp_1)\right)$ is true), it is denoted as a {\it valid} local edge. If a valid local edge was found, the method progresses onto step 2. Threshold $L_C$ is given by $L_C = \max(L,|\hat{\mu}_T-\hat{\mu}_B|/2)$, where $L = 3$ and $C = 0.7$. The overall combination of contextual and local edges using connectivity is summarized in Alg.~\ref{alg:combContextLocalEdges}, for $H(\theta_n) = H$.


\begin{algorithm}[hbt]
   \caption{combination of contextual and local edges using connectivity\label{alg:combContextLocalEdges}}
\begin{algorithmic}[1]
\STATE {\bf input:} directional data $\bnabla_{\!\theta}\bI$, parameters $\hat{\bmu}$ and $\hat{\bsigma}^2$, maximum distance threshold $d$, contextual $C$ and local $L$ thresholds, TV distances $\bDelta(\cdot,\cdot)$
    \STATE $\bE \leftarrow 0$
    \STATE \% for every pixel in the image (assuming $H(\theta_n) = H$)
    \FOR{$y \in \{1,\hdots,S_y\}$}
    \STATE step = 'search for start'
    \FOR{$x \in \{1,\hdots,S_x\}$}
    \STATE \% get contextual edge data
    \STATE $\bp_{1_T} = \bA(x-\left[\sin\theta_n\right],y-\left[\cos\theta_n\right],\theta_n)$
    \STATE $\bp_{1_B} = \bA(x+\left[\sin\theta_n\right],y+\left[\cos\theta_n\right],\theta_n)$
    \STATE $\hat{\mu}_T = \hat{\bmu}(\bp_{1_T})$,  $\hat{\sigma}_T^2 = \hat{\bsigma}^2(\bp_{1_T})$, $\hat{\mu}_B = \hat{\bmu}(\bp_{1_B})$, $\hat{\sigma}_B^2 = \hat{\bsigma}^2(\bp_{1_B})$
    \STATE $\mu'_B = |\hat{\mu}_T - \hat{\mu}_B|/\min(\hat{\sigma}_T,\hat{\sigma}_B)$, $\sigma'_B = \max(\hat{\sigma}_T,\hat{\sigma}_B)/\min(\hat{\sigma}_T,\hat{\sigma}_B)$
    \STATE $\delta_{{X_T}{X_B}} = \bDelta(\mu'_B,\sigma'_B)\text{sgn}(\hat{\mu}_T-\hat{\mu}_B)$
    \STATE $L_C = \max(L,|\hat{\mu}_T-\hat{\mu}_B|/2)$
    \STATE \% combine contextual and local edges
    \STATE {\bf if }step = search for start' {\bf then }Alg.~\ref{alg:combContextLocalEdgesStep12}~{\bf else }Alg.~\ref{alg:combContextLocalEdgesStep3}~{\bf end if}
    \ENDFOR
    \ENDFOR
\STATE {\bf output:} Edge map $\bE$
\end{algorithmic}
\end{algorithm}

\subsubsection{Marking of initial area}


The contextual and valid local edge found in step 1 indicates that a set of connected edge points may start at that location. That occurs if the $M$ pixels between the top and bottom set of samples contain valid local edges that are sufficiently connected to each other. Step 2 verifies this and, if true, marks the pixels as {\it connected edge points}.

\begin{figure}[hbt!]
\centerline{
\includegraphics[width=.6\linewidth]{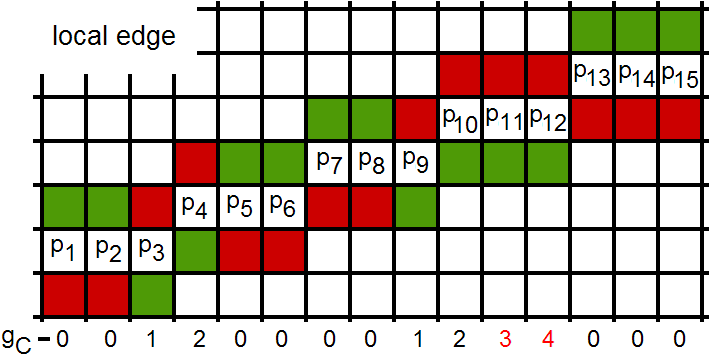}
}
\caption{Illustration of the sign of local edges and gap counter, $g_C$ (using the contextual edge illustrated in Fig.~\ref{fig:threePhases1}). Using $d = 2$ (for illustration), the four consecutive local edges of contrary sign makes step 2 end abruptly while analyzing pixel $\bp_{11}$.}\label{fig:threePhases2}
\end{figure}

Pixels $\bp_m \in \{\bp_1,\hdots,\bp_M\}$ are checked sequentially, as illustrated in Fig.~\ref{fig:threePhases2} (where $\bp_m = \bA(x+m-1,y,\theta_n)$, if $H(\theta_n) = H$). Whenever a valid local edge is found, {\it i.e.}, $\text{sgn}\left(\delta_{{X_T}{X_B}}\right)\bnabla_{\!Q(\theta_n)}(\bp_m) \geq L_C$, gap counter $g_C$ is nulled --- otherwise, $g_C$ is incremented. Note that, while in approaches such as~\cite{ConnectiveHT93}, the distance between binary local edge points is the only criteria to judge whether pixels are connected with each other or not, our approach (as also illustrated in Fig.~\ref{fig:threePhases2}) requires that the sign of the edge points matches the sign of the connected edges. Local edges with opposite signs --- even strong ones --- are discarded in the same way as non-edges.

If the gap counter exceeds the maximum distance threshold $d$, the gap between two valid edge points is too large and step 2 ends unsuccessfully, returning to step 1 for further searching. If the gap counter never exceeded $d$, the pixels are marked as edge points of sign $\bE(\bp_m) = \text{sgn}\left(\delta_{{X_T}{X_B}}\right)$ and the method proceeds to step 3. We allow gaps of up to $d = 5$ pixels in this paper. Step 1 and 2 of the combination of contextual and local edges are summarized in Alg.~\ref{alg:combContextLocalEdgesStep12}, for $H(\theta_n) = H$.


\begin{algorithm}[hbt]
   \caption{find initial set of connected edge points (assumes $H(\theta_n) = H$)\label{alg:combContextLocalEdgesStep12}}
\begin{algorithmic}[1]
\STATE {\bf input:} signed TV distance $\delta_{{X_T}{X_B}}$, directional data $\bnabla_{\!Q(\theta_n)}\bI$, maximum distance threshold $d$, contextual $C$ and local $L_C$ thresholds
    \STATE $\bp_1 = \bA(x,y,\theta_n)$
    \IF{$|\delta_{{X_T}{X_B}}| \geq C$, $|\bnabla_{\!Q(\theta_n)}(\bp_1)| \geq L_C$, $\delta_{{X_T}{X_B}}\bnabla_{\!Q(\theta_n)}(\bp_1) > 0$}
    \STATE \% found initial edge, implement step 2
    \WHILE{$m \in \{1,\hdots,M\}$ and $g_C \leq d$}
     \IF{$\text{sgn}\left(\delta_{{X_T}{X_B}}\right)\bnabla_{\!Q(\theta_n)}(\bp_m = \bA(x+m-1,y,\theta_n)) \geq L_C$}
    \STATE \% found valid local edge
    \STATE $g_C \leftarrow 0$
    \ELSE
    \STATE $g_C \leftarrow g_C + 1$
    \ENDIF
    \ENDWHILE
    \IF{$g_C  \leq d$}
    \STATE \% does not exceed $d$, mark $M$ edge points
    \FOR{$m \in \{1,\hdots,M\}$}
    \STATE $\bE(\bp_m = \bA(x+m-1,y,\theta_n)) = \text{sgn}\left(\delta_{{X_T}{X_B}}\right)$
    \ENDFOR
    \STATE step = 'mark area'
    \ENDIF
    \ENDIF
\end{algorithmic}
\end{algorithm}

\subsubsection{Marking of connected area}


As the set of connected edge points increase, the contextual and local edges to be checked move progressively along the line. For $H(\theta_n) = H$, this occurs by making $x \leftarrow x+1$ and re-computing both types of edges for contextual and local compliance. The contextual edge is computed as before, by accessing the parameters at positions $\bp_{1_T}$ and $\bp_{1_B}$ and computing the new value of $\delta_{{X_T}{X_B}} = \bDelta(\mu'_B,\sigma'_B)\text{sgn}(\hat{\mu}_T-\hat{\mu}_B)$. If $|\delta_{{X_T}{X_B}}| \geq C$ and has the sign of step 1, the contextual edge is valid. 

\begin{figure}[hbt!]
\centerline{
\includegraphics[width=.5\linewidth]{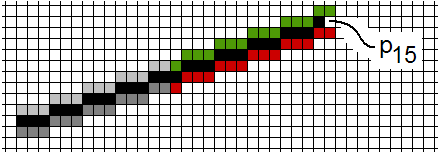}
}
\caption{Illustration of the marking of connected edge points. The connected edge points are marked in black. Contextual and local edge at pixel $\bp_M = \bp_{15}$ are being checked.}\label{fig:threePhases3}
\end{figure}

To determine if there a valid local edge, pixel $\bp_M = \bA(x+M-1,y,\theta_n)$ is checked. If $\text{sgn}\left(\delta_{{X_T}{X_B}}\right)\bnabla_{\!Q(\theta_n)}(\bp_M) \geq L_C$, it a valid local edge and counter $g_C$ is nulled --- otherwise, $g_C$ is incremented. If $g_C \leq d$ and the contextual edge is valid, pixel $\bp_M$ is marked as a connected edge point of sign $\bE(\bp_M) = \text{sgn}\left(\delta_{{X_T}{X_B}}\right)$. Otherwise, the set of connected edge points ended. The last $g_C$ non-connected edge points are unmarked and the method returns to step 1, to search for a new set. This is illustrated in Fig.~\ref{fig:threePhases3} and summarized in Alg.~\ref{alg:combContextLocalEdgesStep3}. \\

\begin{algorithm}[hbt]
   \caption{mark connected edge points (assumes $H(\theta_n) = H$)\label{alg:combContextLocalEdgesStep3}}
\begin{algorithmic}[1]
\STATE {\bf input:} signed TV distance $\delta_{{X_T}{X_B}}$, directional data $\bnabla_{\!Q(\theta_n)}\bI$, maximum distance threshold $d$, contextual $C$ and local $L_C$ edge thresholds
    \IF {$|\delta_{{X_T}{X_B}}| \geq C$}
    \IF{$\text{sgn}\left(\delta_{{X_T}{X_B}}\right)\bnabla_{\!Q(\theta_n)}(\bp_M = \bA(x+M-1,y,\theta_n)) \geq L_C$}
    \STATE \% found valid local edge
    \STATE $g_C \leftarrow 0$
    \ELSE
    \STATE $g_C \leftarrow g_C + 1$
    \ENDIF
    \IF{$g_C  \leq d$}
           \STATE \% does not exceed $d$, mark $\bp_M$ as a connected edge point
           \STATE $\bE(\bp_M) = \text{sgn}\left(\delta_{{X_T}{X_B}}\right)$
    \ELSE
	\STATE \% unmark non-connected edge points
           \FOR {$m \in \{M-g_C+1,\hdots,M\}$}
           \STATE $\bE(\bp_m = \bA(x+m-1,y,\theta_n)) = 0$
           \ENDFOR
          \STATE step =  'search for start'
    \ENDIF
    \ELSE
        \STATE step =  'search for start'
    \ENDIF
\end{algorithmic}
\end{algorithm}

\section{Rectangle fitting}
\label{sec:recFittingValidation}

\begin{figure}[hbt!]
\centerline{
\includegraphics[width=.99\linewidth]{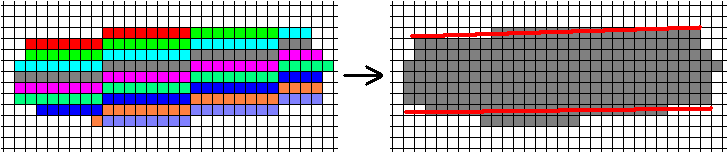}
}
\caption{The multiple connected edges along lines (left), obtained in Section~\ref{sec:combContextLocal}, are joined into a single area and lines are fit to the upper and lower limits (right).}\label{fig:fitLinesToRegion}
\end{figure}

The edge map that is computed for every $\theta_n$ contains positive and negative edge points. Due to the handling of connectivity, the edge points of a line segment do not exhibit gaps between them and, consequently, a simple region growing scheme is sufficient to obtain the areas that form a rectangle. This avoids expensive search mechanisms for overcoming gaps, as the costly scheme used in our previous HT-based method, in~\cite{ourTIPHTPaper12}. This section works by:


\noindent {\it - 1. Region growing and labeling} - A unique identification number is given to each set of connected edge points with the same sign, as illustrated in Fig.~\ref{fig:fitLinesToRegion}.

\noindent {\it - 2. Fitting a rectangle to each area} - Rectangle fitting can occur in various ways (see~\cite{LSD10} for a brief summary). In our method, we start by fitting a line to the upper and lower limits of each area, as illustrated in Fig.~\ref{fig:fitLinesToRegion}. Then, using the average angle of both fitted lines, $\theta_R = (\theta_{upper}+\theta_{lower})/2$, we obtain the start and end of each line segment.

\noindent {\it - 3. Validate rectangles} - A validation step is needed to eliminate non-rectilinear structures in the image. Although various criteria can be used, we require only that the upper and lower fitted lines (see above) should have similar angles, $\theta_{upper}\sim \theta_{lower}$, and the average angle should lie inside the permitted range, $\theta_R \in [\theta_n-180^{\circ}/2N, \theta_n+180^{\circ}/2N]$.

\begin{figure}[hbt!]
\centerline{
\includegraphics[width=.4\linewidth]{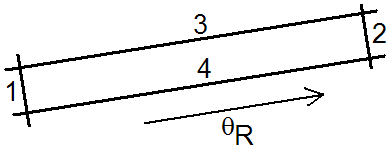}
}
\caption{Rectangle limits.}\label{fig:howRectanglesAreOutputted}
\end{figure}

\noindent {\it - 4. Store rectangles} - Valid rectangles are then stored as angle $\theta_R$ and the parameters corresponding to the four limiting lines, represented in Fig.~\ref{fig:howRectanglesAreOutputted}.

\section{Experiments}
\label{sec:experiments}

We single out demonstrative results of our method, which we contrast with the ones obtained with the standard HT~\cite{DudaHart72}, the state-of-the-art of local methods LSD~\cite{LSD10} (the superiority of LSD when compared to several other local methods is thoroughly demonstrated in~\cite{LSD10}), and our previous HT-based method, denoted STRAIGHT~\cite{ourTIPHTPaper12}. We describe experiments with synthetic images, which help characterize the general behavior of the new method. Then, we present results obtained with several real world images, which demonstrate its performance in practical application. Finally, we discuss the computational complexity of these methods.



\subsection{Synthetic images}

\begin{figure}[htb]
\centerline{
\includegraphics[width=.25\linewidth]{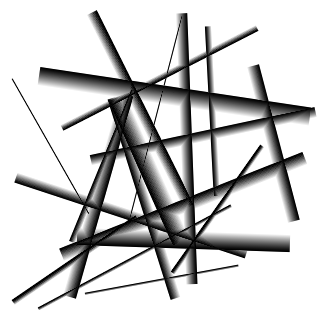}
\includegraphics[width=.25\linewidth]{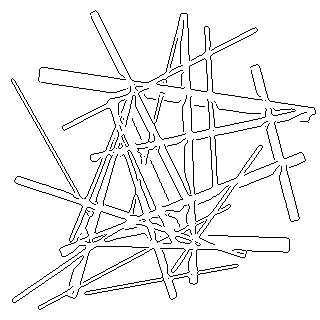}
\includegraphics[width=.25\linewidth]{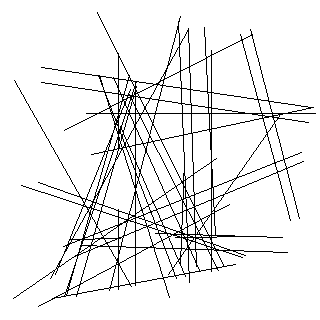}
}
\centerline{
\includegraphics[width=.25\linewidth]{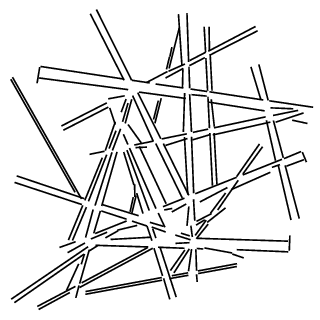}
\includegraphics[width=.25\linewidth]{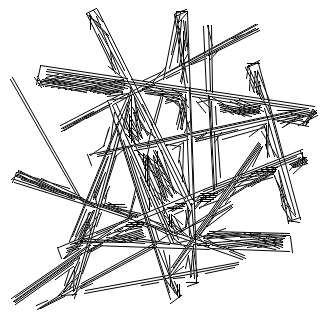}
\includegraphics[width=.25\linewidth]{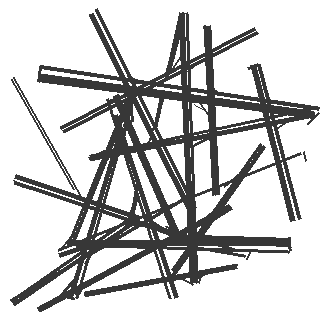}}
\caption{Image with prominent lines. Top left to right: original image, result of Canny edge detector~\cite{canny86}, and standard HT \cite{DudaHart72}. Bottom left to right: LSD~\cite{LSD10}, STRAIGHT~\cite{ourTIPHTPaper12}, and the proposed method.}\label{fig:syntheticImages}
\end{figure}

We start by illustrating the behavior of the algorithms when dealing with an image made up of intersecting line segments of multiple lengths and widths, shown on the top left of Fig.~\ref{fig:syntheticImages}. By comparing the edges computed by the Canny edge detector~\cite{canny86} with the line segments that the HT extracts from them, on the top middle and right of Fig.~\ref{fig:syntheticImages}, respectively, we conclude that the HT succeeds in correctly extracting the lines from this image. This occurs because line segments are long, not in a large number, and the HT does not require connectivity, therefore being able to overcome the multiple line crossings. On the other hand, the results of the LSD method, shown in the bottom left image of Fig.~\ref{fig:syntheticImages}, illustrate that local methods fail to overcome line crossings and splits them. This occurs because local methods require absolute connectivity, {\it i.e.}, that edge points are perfectly chained together. In the particular case of the LSD, the state-of-the-art of local methods, edge points must have approximately constant direction as well. Although the results of STRAIGHT show that it is able to extract thin line segments regardless of the intersections, it is unable to deal with thick ones, originating multiple erroneous detections. Our semi-global method succeeds in extracting line segments of all lengths and widths, with few errors. A pair of twin segments is extracted for each segment in the original image because both light-to-dark and dark-to-light transitions are detected.

\begin{figure}[htb]
\centerline{
\includegraphics[width=.19\linewidth]{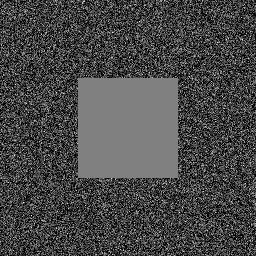}
\includegraphics[width=.19\linewidth]{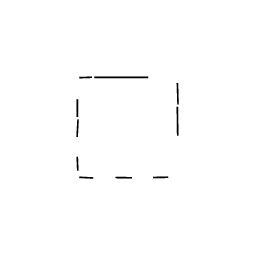}
\includegraphics[width=.19\linewidth]{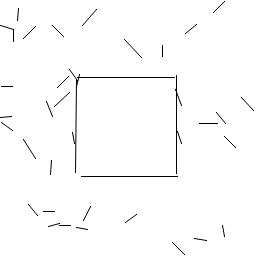}
\includegraphics[width=.19\linewidth]{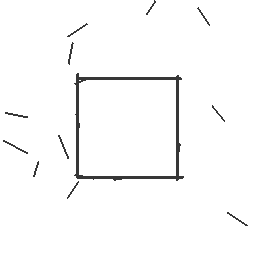}
}
\vspace{0.25cm}
\centerline{
\includegraphics[width=.19\linewidth]{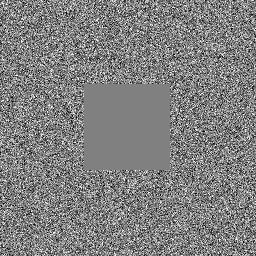}
\includegraphics[width=.19\linewidth]{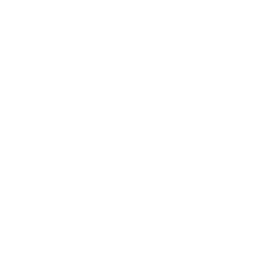}
\includegraphics[width=.19\linewidth]{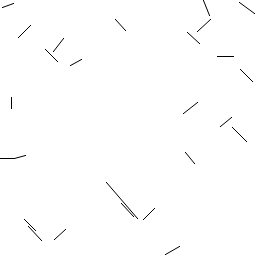}
\includegraphics[width=.19\linewidth]{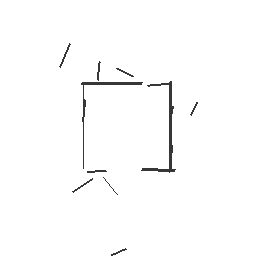}
}
\vspace{0.25cm}
\centerline{
\includegraphics[width=.19\linewidth]{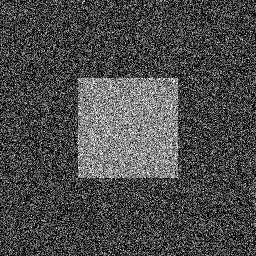}
\includegraphics[width=.19\linewidth]{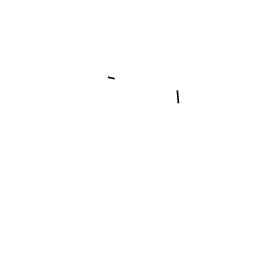}
\includegraphics[width=.19\linewidth]{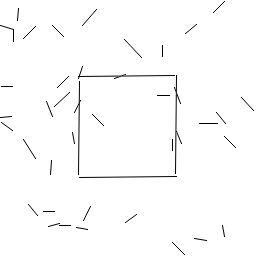}
\includegraphics[width=.19\linewidth]{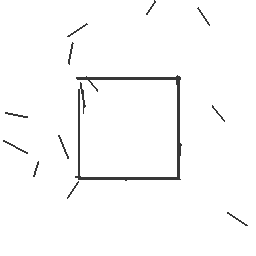}
}
\caption{Textured images. From left to right: original image, result of LSD \cite{LSD10}, STRAIGHT \cite{ourTIPHTPaper12} and the proposed method.}\label{fig:noiseResults}
\end{figure}

We now illustrate the behavior of the algorithms in capturing transitions between differently textured regions. This simulates low signal-to-noise scenarios that occur when using very low thresholds in edge detection, for increased sensibility, where real transitions should be extracted successfully, while avoiding false ones. We use the synthetic images in the left column of Fig.~\ref{fig:noiseResults}, which were generated by adding noise to a piecewise constant map (the top and bottom images were first used in~\cite{ourTIPHTPaper12} and are used here to enable a simple comparison between the methods). The top image represents a simpler scenario, where one of the areas involved in the transition is perfectly smooth. The central image represents the same scenario, except that the mean value of both regions is now equal, making the variance the single discriminating factor. The bottom image simulates two smooth objects in a low signal-to-noise image. Due to the effect of noise, the local edge detection in LSD can not produce edges with constant direction along real segments and the LSD fails to extract line segments, except for parts of the top image. The results of STRAIGHT and the proposed method, in the two rightmost columns of Fig.~\ref{fig:noiseResults}, show that both overcome noise and succeed in extracting the line segments for the top and bottom images (the few short segments correspond to accidental connected alignments in the random texture). By allowing samples with the same mean but different variances to be classified as edges, by using two-sample tests and, in particular, the TV distance, the proposed method is the only one that succeeds in obtaining most of the real segments of the figure in the middle row.

\subsection{Real images}

\begin{figure}[htb!]
\centerline{
\includegraphics[width=.49\linewidth]{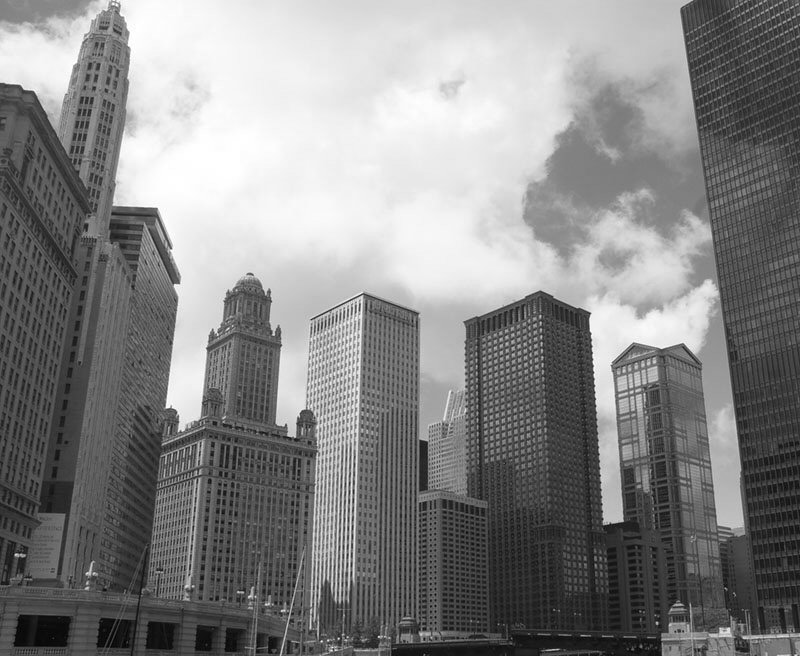}\hspace*{.1cm}
\includegraphics[width=.49\linewidth]{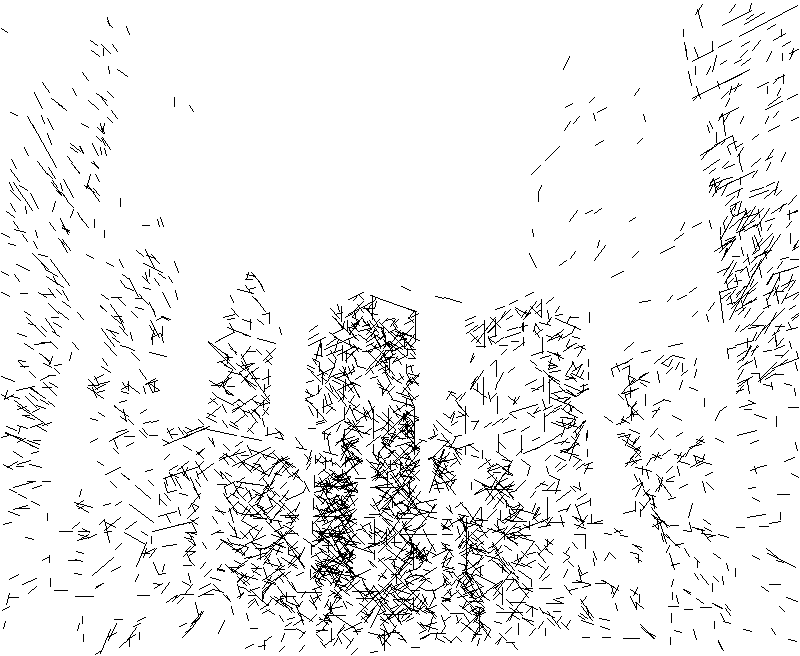}
}\vspace*{.1cm}
\centerline{
\includegraphics[width=.49\linewidth]{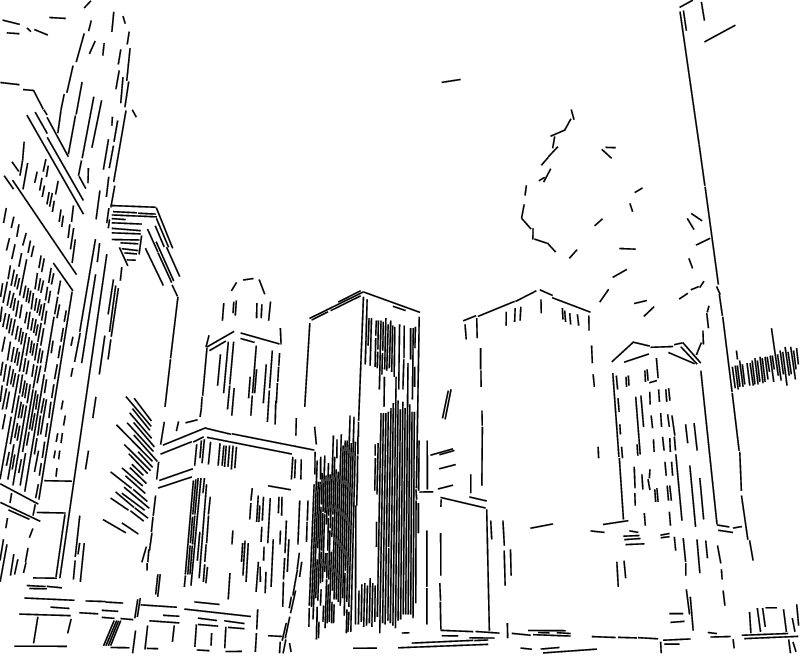}\hspace*{.1cm}
\includegraphics[width=.49\linewidth]{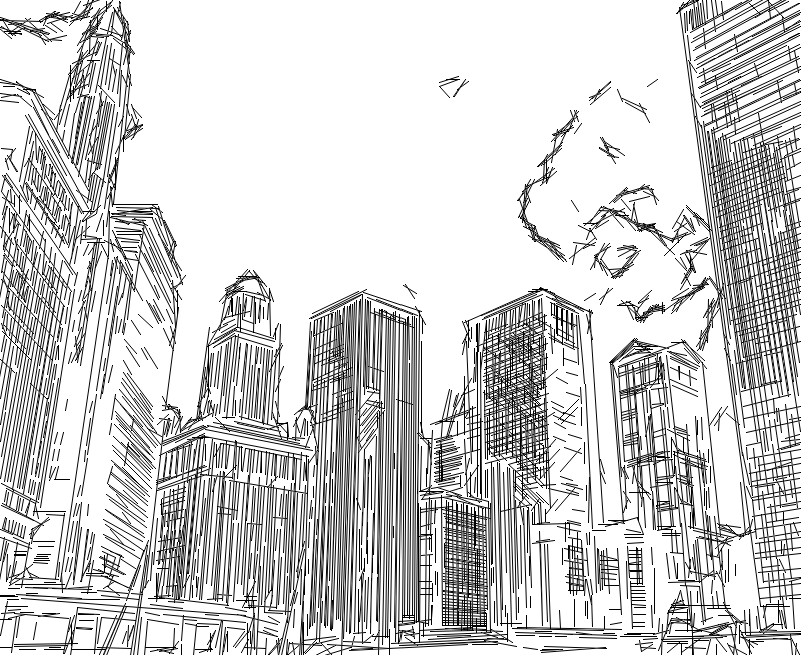}
}\vspace*{.1cm}
\centerline{
\includegraphics[width=.49\linewidth]{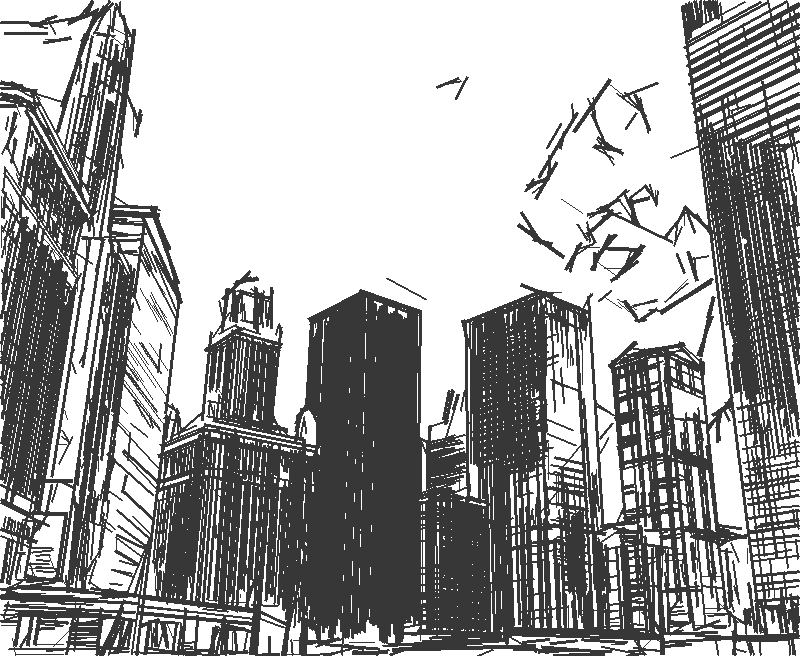}\hspace*{.1cm}
\includegraphics[width=.49\linewidth]{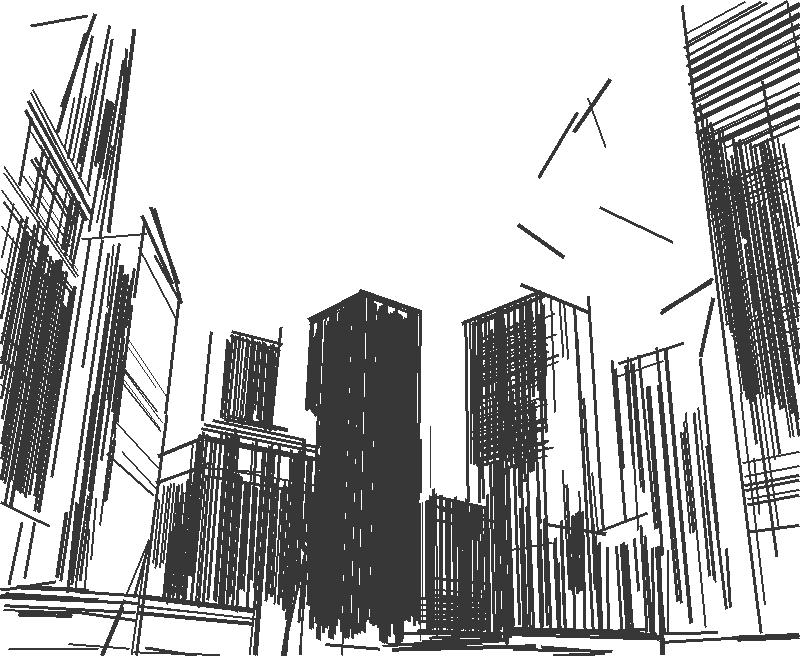}
}
\caption{Top left: image. Top right: HT~\cite{DudaHart72}. Middle left: LSD~\cite{LSD10}. Middle right: STRAIGHT~\cite{ourTIPHTPaper12}. Bottom left: our method. Bottom right: our method (longer segments).}\label{fig:building}
\end{figure}

We start by showing a challenging image that was first used in~\cite{ourTIPHTPaper12} to demonstrate the ability of STRAIGHT in dealing with the dense packing of line segments of multiple lengths that cross each other. In the top right image of Fig.~\ref{fig:building}, we display the results of the HT~\cite{DudaHart72}, showing that extraction fails altogether for not being able to cope with the large number of edge points (this is explained in detail in~\cite{ourTIPHTPaper12}). On the middle left image of Fig.~\ref{fig:building}, we display the results of LSD~\cite{LSD10}, showing that a subset of the line segments are in fact detected. A closer look reveals that those are only the line segments that do not cross other structures and also that several longer segments are detected as fragmented ones. The results of STRAIGHT are shown in the middle right image of Fig.~\ref{fig:building} and we see that it succeeds in extracting the vast majority of the line segments in the image (exceptions are those which exhibit low contrast). Equally good results are obtained by the method we propose, shown in the bottom images of Fig.~\ref{fig:building}, with the difference that STRAIGHT needed 52 seconds to process this image while our semi-global method needed only about 4 seconds on the same machine. The bottom right image displays only the line segments that have length greater than 50 pixels, illustrating that line segments are not fragmented.

%

\begin{figure}[htb]
\centerline{
\includegraphics[width=.49\linewidth]{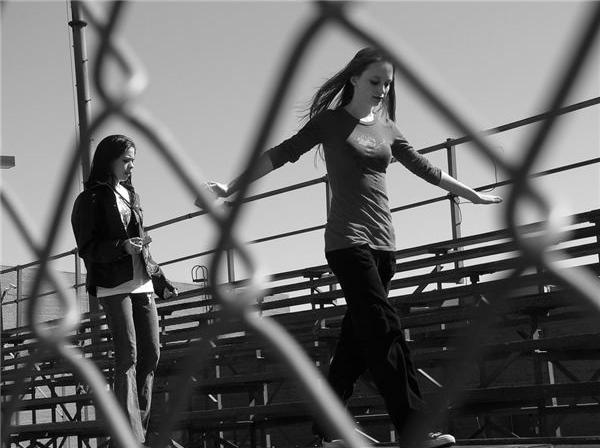}\hspace*{.1cm}
\includegraphics[width=.49\linewidth]{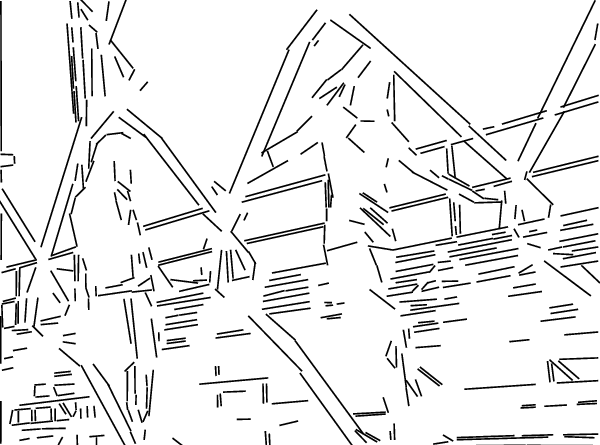}
}\vspace*{.1cm}
\centerline{
\includegraphics[width=.49\linewidth]{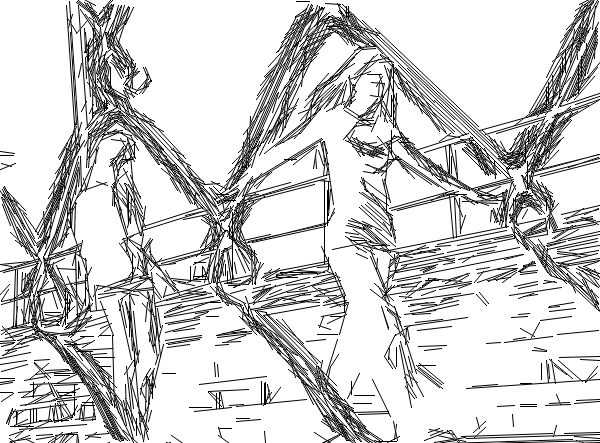}\hspace*{.1cm}
\includegraphics[width=.49\linewidth]{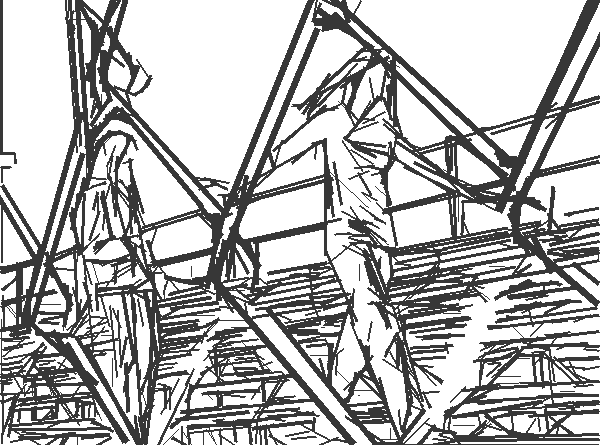}
}
\caption{Top left: image. Top right: LSD~\cite{LSD10}. Bottom left: STRAIGHT~\cite{ourTIPHTPaper12}. Bottom right: our method.}\label{fig:girls}
\end{figure}

Fig.~\ref{fig:girls} presents another illustrative case. It was obtained by processing an image containing a complex scene of line segments (many of which of low contrast) occluded by a net that is large and out-of-focus. The result of LSD~\cite{LSD10} shows that most low contrast segments were not extracted and that others are fragmented in multiple pieces. The fragmentation of line segments is improved in STRAIGHT~\cite{ourTIPHTPaper12} but low contrast segments are equally not extracted and the thick lines of the net originate a multitude of erroneous line segments. On the other hand, our proposed method extracts most line segments, including low contrast and thick ones, with little fragmentation. The extraction of low contrast line segments is enabled by the better handling of variable high frequency variation by two-sample statistical tests. Low contrast and blurred line segments occur often in realistic scenarios and are tackled by our method. 

\begin{figure}[htb]
\begin{minipage}[b]{0.48\linewidth}
\centerline{
\includegraphics[width=.49\linewidth]{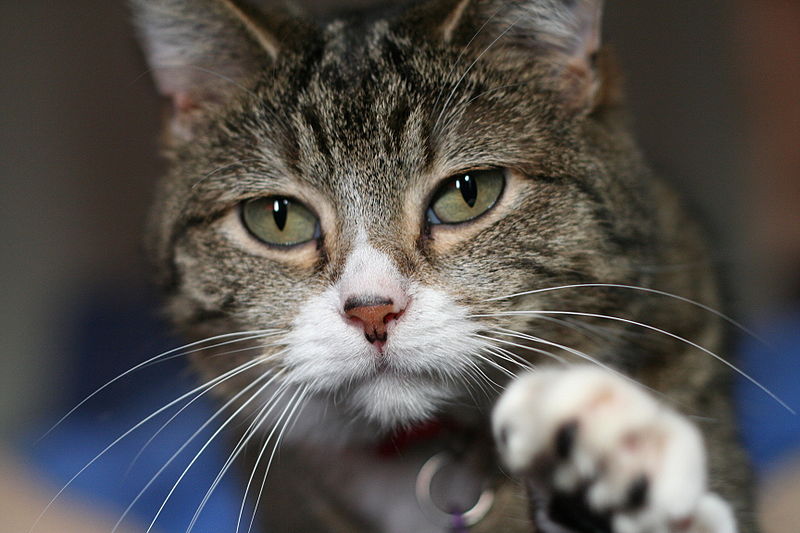}\hspace*{.05cm}
\includegraphics[width=.49\linewidth]{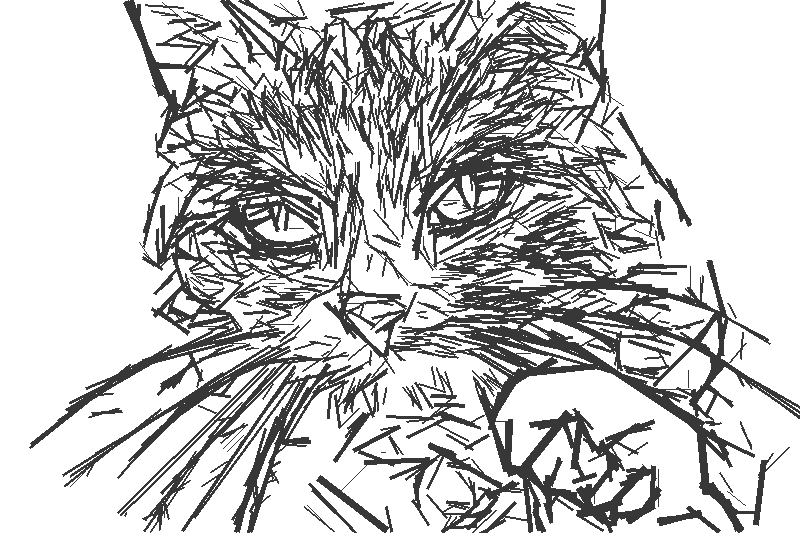}}\vspace*{.1cm}
\centerline{
\includegraphics[width=.49\linewidth]{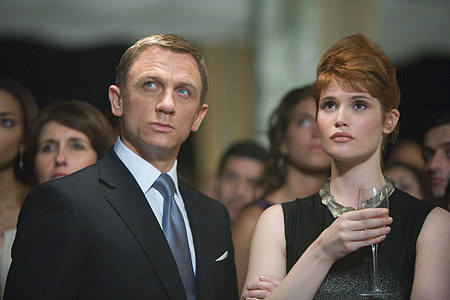}\hspace*{.05cm}
\includegraphics[width=.49\linewidth]{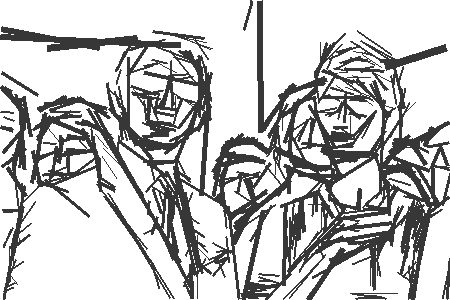}}
\end{minipage}
\begin{minipage}[b]{0.48\linewidth}
\centerline{
\includegraphics[width=.49\linewidth]{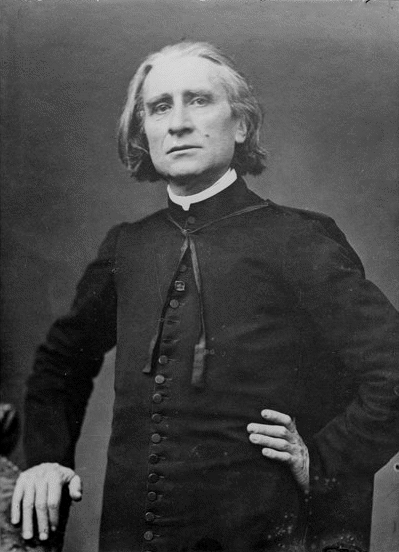}\hspace*{.05cm}
\includegraphics[width=.49\linewidth]{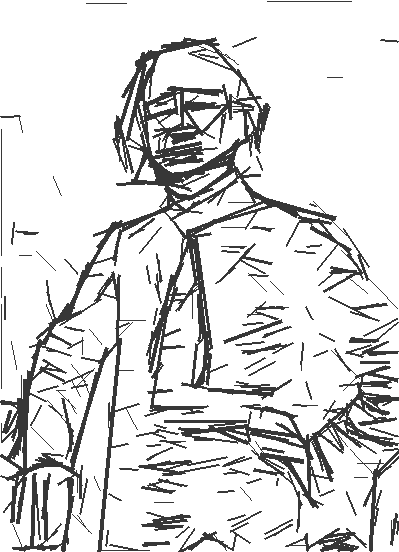}}
\end{minipage}\vspace*{.1cm}
\centerline{
\includegraphics[width=1.0\linewidth]{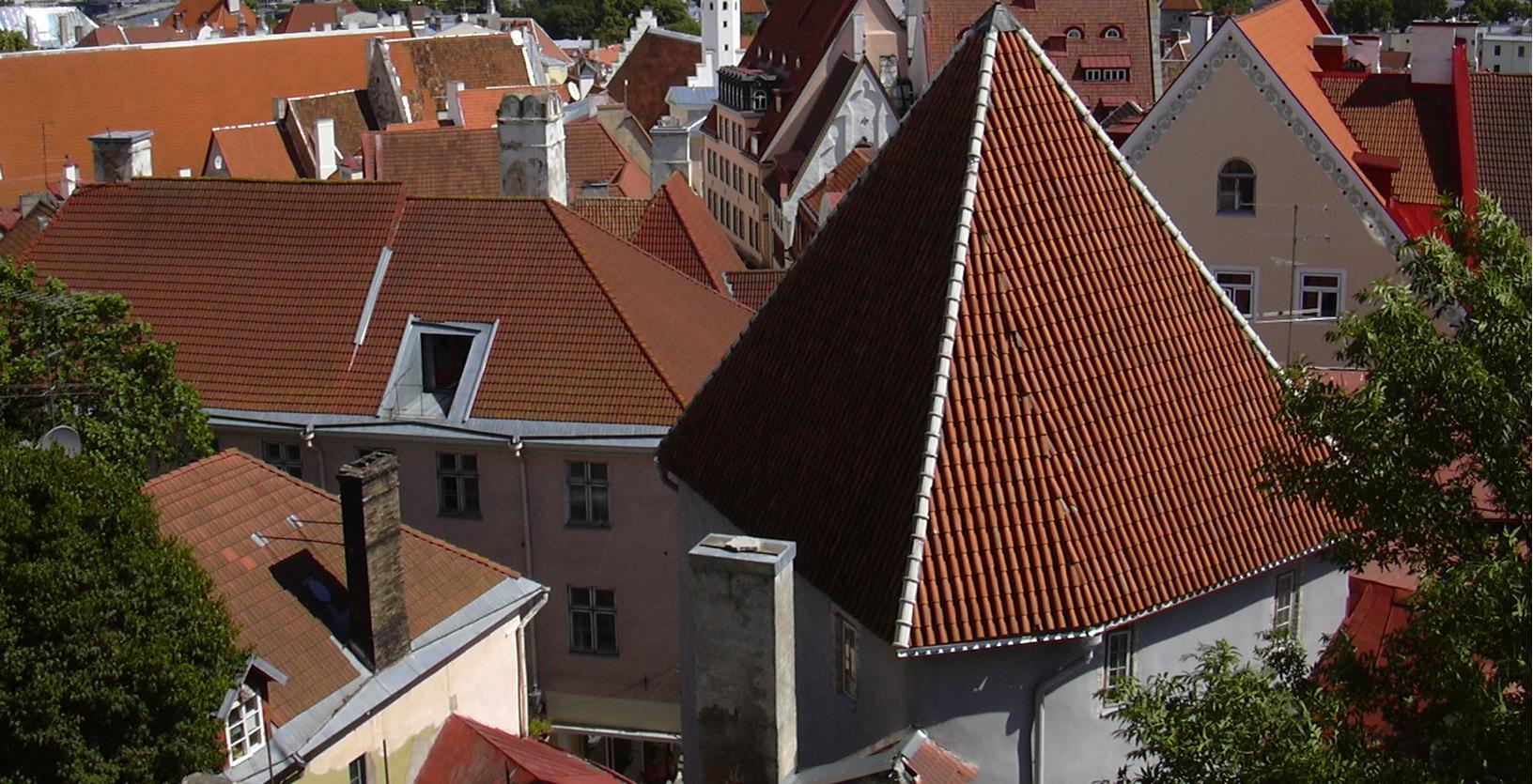}}\vspace*{.1cm}
\centerline{
\includegraphics[width=1.0\linewidth]{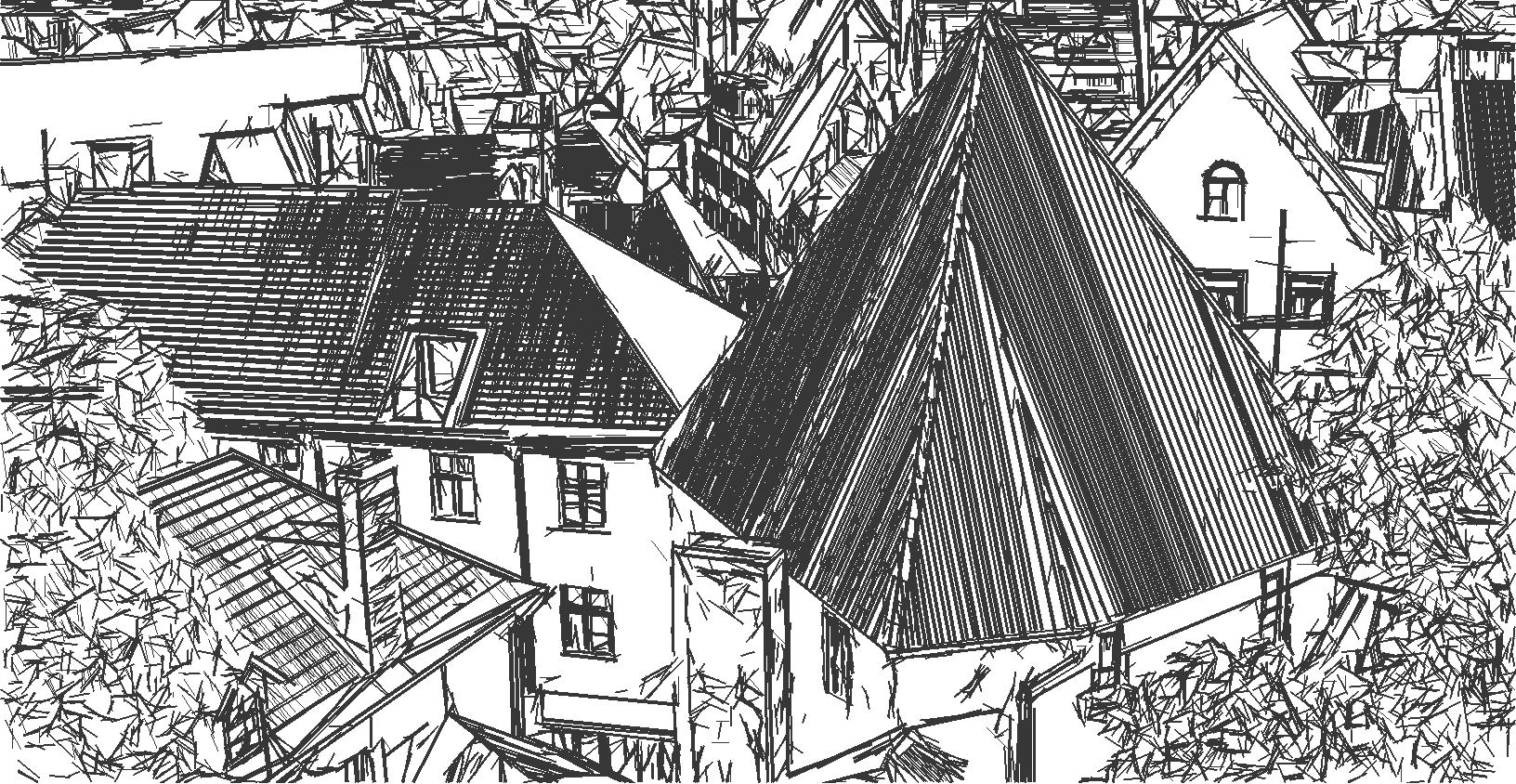}}
\caption{Results of our method for several kinds of real images.}\label{fig:manyResults}
\end{figure}

Finally, Fig.~\ref{fig:manyResults} presents results of using our method with real images of various kinds. As desired, the vast majority of long line segments are extracted without artificial fragmentation, despite the multiple segment crossings. Also note that, although some of these images have edges that form curves, our method succeeds in approximating these sections in a piecewise linear way, {\it i.e.}, by a sequence of rectilinear line segments.

\subsection{Computational complexity}

\begin{figure}[htb]
\centerline{
\includegraphics[width=.49\linewidth]{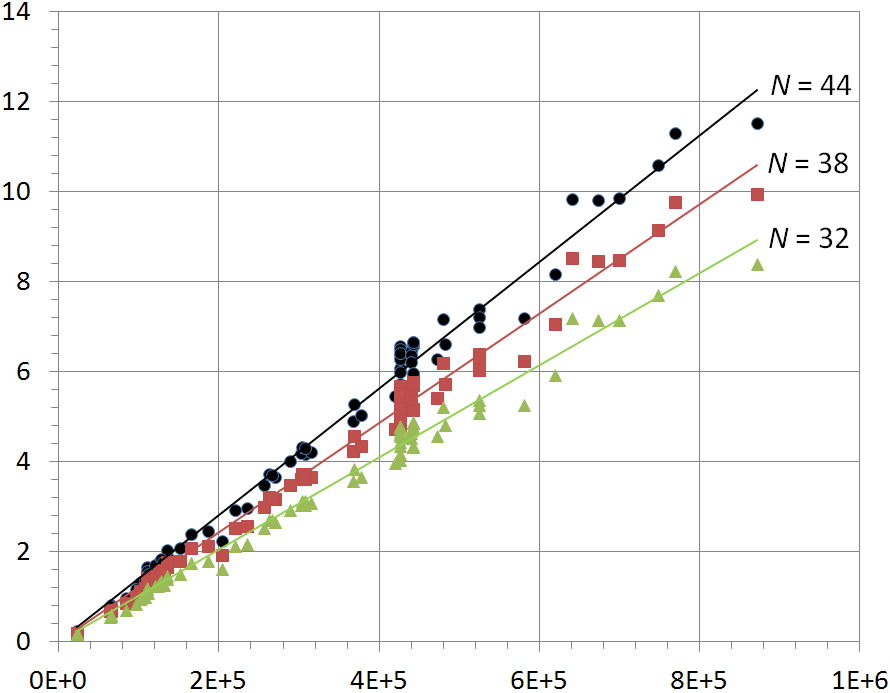}
\includegraphics[width=.49\linewidth]{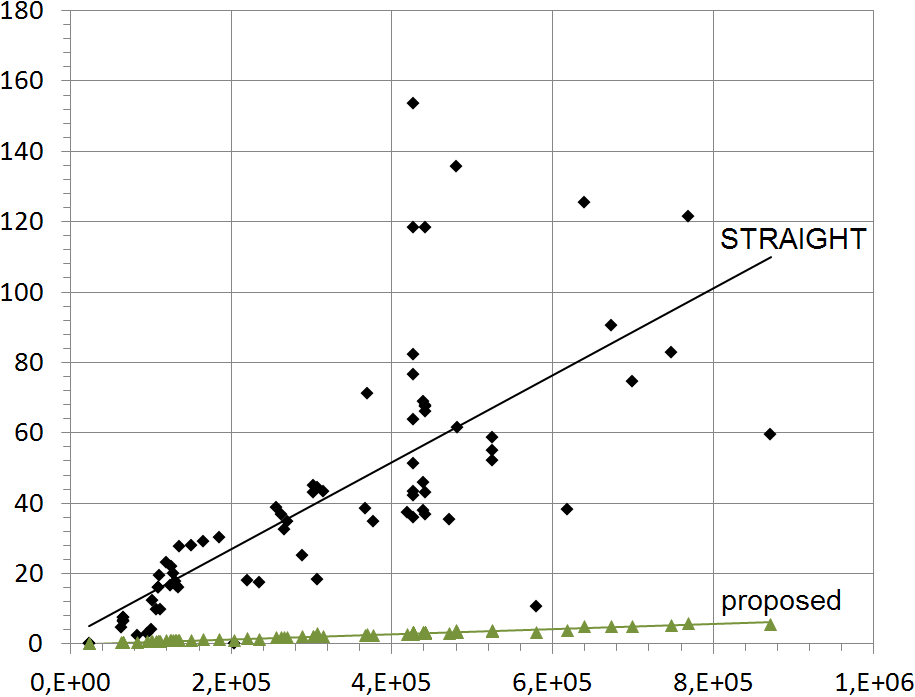}
}
\caption{Extraction times (in seconds) as a function of pixel count. Left: proposed method, for $N$ number of directions. About 71K pixels per second are processed using $N=44$ directions, 82K pixels/s using $N=38$, and 97K pixels/s using $N=32$. Right: proposed method versus STRAIGHT, with $N=32$. STRAIGHT processes roughly 8K pixels/s.}\label{fig:computationTime}
\end{figure}

The most computationally intensive portion of our method is the calculation of contextual and local edges and their combination into continuous edge maps. By using a running average framework, the calculation of contextual edges depends only linearly on the pixel count and the number of directions, $N$. Such linear dependency also occurs in the calculation of local edges and in their combination with contextual ones. This is confirmed in the left of Fig.~\ref{fig:computationTime}, which shows the computation time\footnote{All experiments were performed on an $\text{Intel}^{\copyright}$ 2.67 GHz machine, with all methods implemented in optimized C\# code, except where noted.} needed by the proposed method to extract line segments for multiple images, as a function of pixel count and the number of directions that are used, $N$. The trend line for $N=32$, the number of directions that are used in all experiments, indicates that about 97K pixels are processed at each second ({\it e.g.}, the line segments in the $512\times 512$ Lenna image are extracted in about $2.7$ seconds). 

The complexity of STRAIGHT~\cite{ourTIPHTPaper12} increases linearly with the number of edge points, as the dominating factor in the calculations is the updating of the Hough space of each local HT. In our experiments, the computation time of STRAIGHT was always considerably higher than the proposed method ({\it e.g.}, the line segments in the Lenna image are extracted in about $36.8$ seconds), which is confirmed on the right of Fig.~\ref{fig:computationTime}. This figure shows the computation time needed by both methods to extract line segments for multiple images, as a function of pixel count and for $N=32$ directions. Although the performance of STRAIGHT varies non-linearly with the pixel count, the rough trendline that is fitted helps in illustrating the significant computational advantage of the method we propose of about one order of magnitude.

The standard HT requires filling an accumulator array, which depends linearly on the total number of edge points. In our experiments, the standard HT required approximately 1 to 10~seconds to process each image in this paper, using $\text{MATLAB}^{\copyright}$ code. 

Reference~\cite{LSD10} states that the complexity of the LSD method depends linearly with the image pixel count, as illustrated by a plot therein showing the calculation time needed to extract line segments in various images. In the worst case scenario, {\it i.e.}, images made up of noise, LSD is able to process about 240K pixels per second. Although the main advantage of local methods such as the LSD is its low computational complexity, with the drawback of only dealing successfully with simple scenarios, the amount of pixels processed by the LSD is only about 2--3 times greater than our proposed method.

Although only STRAIGHT and our proposed method can deal with the complex images that arise in practice, the results above show that the computational complexity of our method is significantly below STRAIGHT. Furthermore, the complexity of our method is comparable with local methods, {\it i.e.}, it is only about 2--3 times more complex than the LSD method, despite the ability to handle complex scenarios. This indicates that our method is efficient in extracting segments of all lengths and widths in complex scenarios.

\section{Conclusion}
\label{sec:conclusions}

We have presented a new semi-global method for line segment extraction. Our method combines contextual and local edges, with explicit handling of connectivity. Our experiments show that it outperforms current methods for line segment extraction in challenging situations, {\it e.g.}, when dealing with complex images containing several crossing segments of multiple widths, and that its computational efficiency is comparable with simple local methods. We use a contextual edge detection scheme based on two-sample statistical tests, which is a robust way to handle noise.





\bibliographystyle{elsarticle-num-names}
\bibliography{references}







\end{document}